\documentclass[3p]{elsarticle}

\usepackage{hyperref}
\usepackage[normalem]{ulem}
\usepackage{amsmath,bm}
\usepackage{amssymb}
\usepackage{amsfonts}
\usepackage{mathtools}
\usepackage{amsthm}
\usepackage{physics}
\usepackage{graphicx}
\usepackage{epstopdf}
\usepackage{booktabs}
\usepackage{array}
\usepackage{multicol}
\usepackage{multirow}
\usepackage{arydshln}
\usepackage[export]{adjustbox}
\usepackage{pifont}
\usepackage{caption}
\usepackage{subfigure}
\usepackage[table]{xcolor}
\usepackage{nicefrac}
\usepackage{microtype}
\usepackage{url}
\usepackage{wrapfig}
\usepackage{float}
\usepackage{tikz}
\usepackage{tikz-cd}
\usetikzlibrary{shapes, arrows.meta, positioning, calc, fit}
\usepackage[capitalize,noabbrev]{cleveref}
\usepackage{algorithm}
\usepackage{algorithmic}
\usepackage{pdfpages}

\theoremstyle{plain}

\theoremstyle{definition}

\theoremstyle{remark}

\journal{Computer Methods in Applied Mechanics and Engineering}

\begin{document}

\begin{frontmatter}

\title{MENO: MeanFlow-Enhanced Neural Operators for Dynamical Systems}

\author[ccs]{Tianyue Yang}
\author[ccs]{Xiao Xue\corref{cor1}}
\ead{x.xue@ucl.ac.uk}
\cortext[cor1]{Corresponding author}

\address[ccs]{Centre for Computational Science, University College London, London, United Kingdom}

\begin{abstract}
Neural operators have emerged as powerful surrogates for dynamical systems due to their grid-invariant properties and computational efficiency. However, Fourier-based variants inherently truncate high-frequency components in spectral space, resulting in the loss of small-scale structures and degraded prediction quality at high resolutions when trained on low-resolution data.  While diffusion-based enhancement methods can recover multi-scale features, they introduce substantial inference overhead that undermines the efficiency advantage of neural operators. In this work, we introduce MeanFlow-Enhanced Neural Operators (MENO), a novel framework that achieves accurate all-scale predictions with minimal inference cost. By leveraging the improved MeanFlow method, MENO restores both small-scale details and large-scale dynamics with superior physical fidelity and statistical accuracy. We evaluate MENO on three challenging dynamical systems, including phase-field dynamics, 2D Kolmogorov flow, and active matter dynamics, at resolutions up to 256$\times$256. Across all benchmarks, MENO improves the power spectrum density accuracy by up to a factor of 2 compared to baseline neural operators while achieving up to $14\times$ faster inference than the state-of-the-art Denoising Diffusion Implicit Model (DDIM)-enhanced counterparts, effectively bridging the gap between accuracy and efficiency. The flexibility and efficiency of MENO position it as an efficient surrogate model for scientific machine learning applications where both statistical integrity and computational efficiency are paramount.
\end{abstract}


\end{frontmatter}

\section{Introduction}

The accurate and efficient simulation of complex dynamical systems, such as those governing fluid flow, weather patterns, and material science, remains a grand challenge in computational science. These systems are often described by complex, non-linear Partial Differential Equations (PDEs)~\cite{evans2022partial}, such as the Navier-Stokes equations, whose solutions exhibit rich multi-scale structures spanning a wide range of spatial and temporal scales. In particular, small-scale dynamics play a critical role in determining macroscopic behavior, governing energy transfer, dissipation, and long-term system evolution~\cite{pope2001turbulent}. While traditional numerical methods such as Direct Numerical Simulation (DNS)~\cite{lee2015direct} can resolve these small-scale features with high fidelity, they are notoriously expensive, incurring computational costs that are often prohibitive for many practical applications. This has led to the development of approximation methods such as Reynolds-Averaged Navier-Stokes (RANS)~\cite{alfonsi2009reynolds} and Large Eddy Simulation (LES)~\cite{piomelli1999large}, which reduce the computational burden by partially modelling or filtering fine-scale dynamics, at the expense of physical fidelity.

A new paradigm has emerged in recent years with the application of deep learning to these problems. Among the most promising of these are Neural Operators (NOs), a class of models designed to learn mappings between infinite-dimensional function spaces~\cite{li2020fourier,rahman2022u,cao2024laplace, lu2021learning}. Unlike traditional neural networks that operate on finite-dimensional vectors, NOs can learn the underlying solution operator of a PDE. This gives them a powerful property: resolution invariance, which allows them to be trained on low-resolution data and evaluated at higher resolutions, offering a flexible and efficient alternative to traditional numerical solvers. Pioneering architectures such as the Fourier Neural Operator (FNO)~\cite{li2020fourier}, DeepONet~\cite{lu2021learning}, U-shaped Neural Operator (UNO)~\cite{rahman2022u} and the Laplace Neural Operator~\cite{cao2024laplace} have demonstrated remarkable success in learning complex dynamics. Despite their strong theoretical foundations and their frequent characterization as resolution-independent models, the empirical accuracy of neural operators tends to deteriorate as the evaluation resolution increases beyond the training regime. Such resolution-dependent degradation fundamentally limits their applicability to high-fidelity, fine-scale simulations. The root cause of this issue can often be traced to architectural design choices: for example, Fourier-based neural operators rely on a truncated spectral representation, which inherently limits the bandwidth of resolvable modes and leads to the systematic loss of high-frequency components. As a consequence, fine-scale structures essential for accurately capturing multi-scale dynamics are poorly represented in the predicted solutions~\cite{qin2024toward,gao2025discretization,khodakarami2025mitigating}.

To improve the physical fidelity of predictions, generative models provide a promising direction due to their ability to capture fine-scale distribution structures~\cite{dhariwal2021diffusion}. Diffusion-based methods, such as Denoising Diffusion Probabilistic Models (DDPMs)~\cite{ho2020denoising} and DDIMs~\cite{song2020denoising}, have demonstrated strong performance across a wide range of applications~\cite{chihaoui2024zeroshot,chihaoui2024bird,kawar2022ddrm,wang2023sinsr,yue2025arbitrarystepsimagesuperresolutiondiffusion,choi2021ilvr,mokady2022nulltextinversioneditingreal}. However, their practical deployment is often limited by the substantial computational cost associated with multi-step sampling. Related frameworks based on the probability-flow ordinary differential equation (ODE)~\cite{albergo2023stochastic, song2021score}, including Flow Matching (FM)~\cite{lipman2022flow} and Stochastic Interpolants~\cite{albergo2023stochastic}, encounter similar efficiency bottlenecks. These limitations have motivated the development of ``fast-forward'' generative models that enable single-step generation. Representative examples include distillation-based consistency models~\cite{song2023consistency} and recent training-from-scratch approaches, such as Shortcut Models~\cite{frans2024one}, Inductive Moment Matching~\cite{zhou2025inductive}, and Consistency Training~\cite{song2023consistency,lu2024simplifying}. Among these methods, MeanFlow (MF)~\cite{geng2025mean} has emerged as a particularly effective framework by modeling the time-averaged velocity, thereby enabling direct one-step synthesis. Its recent extension, improved MeanFlow (i-MF)~\cite{geng2025improved}, further improves training stability and generative performance, achieving state-of-the-art results without relying on distillation.

In this work, we exploit the efficiency of i-MF to address the limitations of neural operators while retaining their differentiability and computational efficiency. We propose MENO, a two-stage hybrid framework designed to deliver accurate all-scale predictions with minimal inference overhead. Specifically, MENO decouples the prediction task into two complementary components: a standard neural operator first learns the coarse-scale system dynamics, while a downstream generative decoder, constructed based on the i-MF model, recovers the multi-scale predictive details in a single efficient step. The main contributions of this work are:
\begin{itemize}
    \item A novel two-stage framework, MENO, that combines a neural operator with a stochastic generative decoder to achieve accurate and efficient multi-scale predictions of dynamical systems.
    \item The first use of an improved MeanFlow model for one-step generative refinement in scientific machine learning, which is significantly more efficient than diffusion-based counterparts.
    \item We provide comprehensive empirical results demonstrating improved prediction accuracy and faithful statistical property recovery on 3 different cross-physics and high-resolution dynamical systems governed by diverse governing equations.
\end{itemize}
We demonstrate the effectiveness and flexibility of the MENO framework through extensive experimental validation on three benchmark systems, including phase-field dynamics, the two-dimensional Kolmogorov flow, and a two-dimensional active matter system with resolutions up to $256 \times 256$ derived from \textit{The Well} scientific machine learning dataset~\cite{ohana2024well}. MENO achieves high-fidelity predictions for high-resolution dynamical systems with minimal computational overhead. We compare MENO against a widely used approach for generative fidelity refinement of physical fields, namely diffusion model (DM) based enhancement~\cite{shu2023physics, oommen2024integrating}. MENO achieves up to a $14\times$ inference speedup over DM-enhanced counterparts while preserving small-scale accuracy, and it consistently outperforms neural operator baselines in predictive fidelity.

\section{Method}

\subsection{Problem setup}\label{subsec:problem_setup}
We consider the prediction of spatiotemporal fields generated by PDE-governed
dynamical systems. Let $\mathbf{x}_k \in \mathbb{R}^{H \times W \times C}$
denote the high-resolution state of the system at time step $k$, where $H$ and
$W$ are the spatial resolution and $C$ is the number of physical channels.
The corresponding low-resolution state is
$\mathbf{a}_k^{\mathrm{LR}} \in \mathbb{R}^{h \times w \times C}$ with
$h < H,\; w < W$, obtained by applying a spatial coarsening operator to
$\mathbf{x}_k$. The training data consist of paired low- and
high-resolution trajectories
$\mathcal{D}^{\mathrm{LR}}=\{(\mathbf{a}^{\mathrm{LR},j}_0,\ldots,\mathbf{a}^{\mathrm{LR},j}_T)\}_{j=1}^{N}$
and
$\mathcal{D}^{\mathrm{HR}}=\{(\mathbf{x}^j_0,\ldots,\mathbf{x}^j_T)\}_{j=1}^{N}$,
where $T$ is the rollout horizon and $N$ is the number of trajectories.

The objective is to construct an efficient surrogate that, given a
low-resolution initial condition $\mathbf{a}^{\mathrm{LR}}_0$, predicts a
sequence of high-resolution future states
$\{\hat{\mathbf{x}}_1,\ldots,\hat{\mathbf{x}}_T\}$ with
$\hat{\mathbf{x}}_k \in \mathbb{R}^{H \times W \times C}$. Instead of
directly learning the high-resolution time-evolution operator, we
decompose the task into two stages. First, a neural operator
$\mathcal{G}_{\phi}$ is trained to approximate the low-resolution flow map,
\begin{equation}
    \tilde{\mathbf{a}}^{\mathrm{LR}}_{k}
    =
    \mathcal{G}_{\phi}\!\left(
    \tilde{\mathbf{a}}^{\mathrm{LR}}_{k-1}
    \right),
    \qquad
    \tilde{\mathbf{a}}^{\mathrm{LR}}_0 = \mathbf{a}^{\mathrm{LR}}_0,
    \label{eq:no_flow_map}
\end{equation}
producing an autoregressive low-resolution rollout
$\{\tilde{\mathbf{a}}^{\mathrm{LR}}_k\}_{k=1}^{T}$ at low computational cost.

Second, a MeanFlow decoder $u_{\theta}$ maps the low-resolution rollout to
the high-resolution space. Let $U(\cdot)$ denote a fixed upsampling operator
from the low-resolution grid to the high-resolution grid. At inference time,
the upsampled prediction $U(\tilde{\mathbf{a}}^{\mathrm{LR}}_k)$ defines an
intermediate state
\begin{equation}
    \mathbf{z}_k
    = (1-\tau)\,U(\tilde{\mathbf{a}}^{\mathrm{LR}}_k)
    + \tau\,\boldsymbol{\epsilon}_k,
    \qquad
    \boldsymbol{\epsilon}_k \sim \mathcal{N}(0,I),
    \label{eq:intermediate_state}
\end{equation}
where $\tau \in (0,1]$ controls the noise level. The high-resolution
prediction is then obtained by one MeanFlow refinement step,
\begin{equation}
    \hat{\mathbf{x}}_k
    = \mathbf{z}_k - \tau\,\mathbf{u}_{\theta}(\mathbf{z}_k,0,\tau).
    \label{eq:mf_refinement}
\end{equation}
Thus, the neural operator provides the coarse temporal evolution, while the
MeanFlow decoder restores high-resolution spatial structures in a single
generative refinement step.

\subsection{Neural operator for coarse-grid dynamics}\label{subsec:NO-framework}

A neural operator $\mathcal{G}_\theta:\mathcal{A}\to\mathcal{U}$ approximates a target solution operator $\mathcal{G}^\dagger$ between separable Banach function spaces $\mathcal{A}$ and $\mathcal{U}$~\cite{li2020fourier}, and is trained on paired data $\{(\mathbf{a}_j,\mathbf{u}_j)\}_{j=1}^{N}$ with $\mathbf{u}_j=\mathcal{G}^\dagger(\mathbf{a}_j)$ by minimising the empirical risk $N^{-1}\sum_j\mathcal{L}(\mathcal{G}_\theta(\mathbf{a}_j),\mathbf{u}_j)$. Practical architectures, including FNO~\cite{li2020fourier}, UNO~\cite{rahman2022u}, DeepONet~\cite{lu2021learning}, and the Laplace Neural Operator~\cite{cao2024laplace}, share the composition
\begin{equation}
\mathcal{G}_\theta = \mathcal{Q}\circ\mathcal{K}^{(L)}_\theta\circ\cdots\circ\mathcal{K}^{(1)}_\theta\circ\mathcal{P},
\label{eq:no_composition}
\end{equation}
of a lifting layer $\mathcal{P}$, kernel integration layers $\mathcal{K}^{(l)}_\theta$, and a projection layer $\mathcal{Q}$, and we refer the reader to~\cite{li2020fourier} for the standard FNO formulation. The relevant property for our purpose is that, despite their nominal resolution invariance, neural operators degrade markedly when evaluated above their training resolution, with fine-scale structures and statistical properties poorly reproduced in fluid-dynamics settings~\cite{qin2024toward,gao2025discretization,khodakarami2025mitigating}. This motivates the second component of MENO: a generative decoder that restores the high-frequency content lost by the operator backbone.

\subsection{MeanFlow model for fine-scale recovery}
MeanFlow is a one-step generative modeling framework built upon Flow Matching~\cite{lipman2022flow}, which learns the transport velocity between two distributions through a probability flow ODE. In contrast to conventional Flow Matching, MeanFlow models the average transport velocity over a time interval, thereby enabling efficient one-step generation~\cite{geng2025mean}. The details of these two frameworks are discussed below.

\subsubsection{Flow Matching}
Flow Matching is a class of generative models that construct a continuous-time flow between a simple base distribution \(p_{\text{base}}\) (e.g., a Gaussian) and a complex target data distribution \(p_{\text{target}}\)~\cite{lipman2022flow}. This flow is defined by a time-dependent probability density path \(p_t(\mathbf{z}_t)\) and a corresponding velocity field \(\mathbf{v}_t(\mathbf{z}_t)\) that generates it.

A common and tractable approach is to define the flow via a conditional path. Given a data sample \(\mathbf{x} \sim p_{\text{target}}\) and noise \(\boldsymbol{\epsilon} \sim p_{\text{base}}\), the conditional flow is often formulated as the linear interpolant
\begin{equation}
\mathbf{z}_t = a_t \mathbf{x} + b_t \boldsymbol{\epsilon},
\end{equation}
where \(a_t, b_t:[0,1] \rightarrow \mathbb{R}\) are differentiable time-dependent functions. The corresponding conditional velocity field is
\begin{equation}
\mathbf{v}(\mathbf{z}_t | \mathbf{x}, \boldsymbol{\epsilon}) = \frac{\mathrm{d} \mathbf{z}_t}{\mathrm{d}t} = a'_t \mathbf{x} + b'_t \boldsymbol{\epsilon}.
\end{equation}
In this work, we focus on the optimal transport path, defined by \(a_t = 1 - t\) and \(b_t = t\). This simplifies the conditional velocity to \(\mathbf{v}(\mathbf{z}_t | \mathbf{x}, \boldsymbol{\epsilon}) = \boldsymbol{\epsilon} - \mathbf{x}\), establishing a straight-line trajectory between the data and noise in latent space.

The ideal Flow Matching objective targets the marginal velocity that generates $p_t(\mathbf{z}_t)$, but this marginal is intractable. The Conditional Flow Matching (CFM) objective circumvents this by matching the conditional velocity instead, with the same gradient as the ideal loss~\cite{lipman2022flow}:
\begin{equation}
\mathcal{L}_{\text{CFM}} = \mathbb{E}_{t, \mathbf{x}, \boldsymbol{\epsilon}} \| \mathbf{v}_\theta(\mathbf{z}_t, t) - \mathbf{v}(\mathbf{z}_t | \mathbf{x}, \boldsymbol{\epsilon}) \|^2.
\end{equation}
Once $\mathbf{v}_\theta$ is trained, samples are generated by solving the ODE $\mathrm{d}\mathbf{z}_t/\mathrm{d}t = \mathbf{v}_\theta(\mathbf{z}_t, t)$ from $t=1$ (noise) to $t=0$ (data).

\begin{figure*}[t]
    \centering    \includegraphics[width=0.96\linewidth]{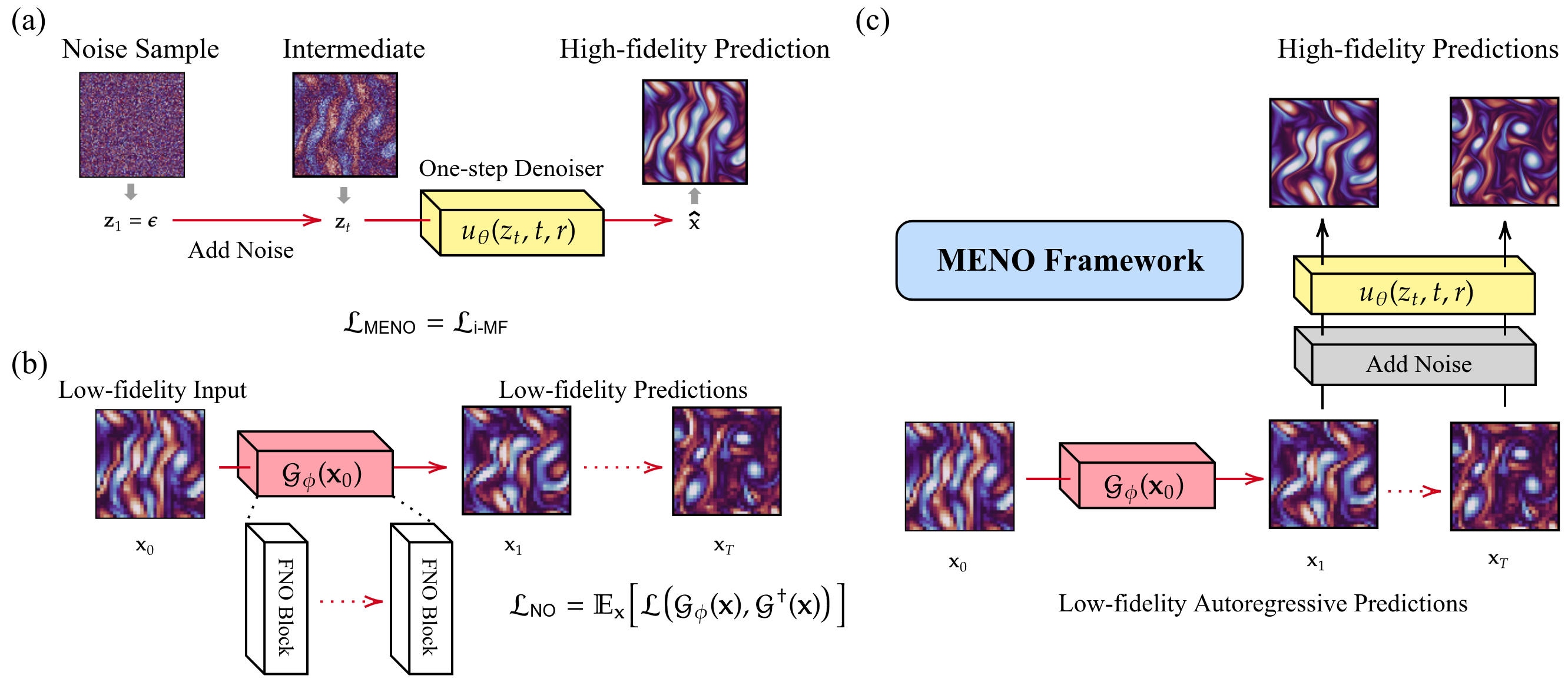}
    \caption{The MeanFlow-Enhanced Neural Operators framework. Panel (a) illustrates training the MeanFlow decoder on high-resolution fields by learning denoising trajectories. Panel (b) shows training an autoregressive neural operator on low-resolution data. Panel (c) combines both components into the full MENO pipeline: given a low-resolution initial condition, the neural operator produces a low-resolution rollout, which is then decoded into a high-resolution prediction by the MeanFlow decoder.}
    \label{fig:MENO_framework}
    \vspace{-1em}
\end{figure*}

\subsubsection{MeanFlow}

While solving the ODE yields accurate samples, it requires multiple evaluations of \(\mathbf{v}_\theta\). The MeanFlow model improves sampling efficiency by directly modeling the average velocity over a time interval~\cite{geng2025mean}. For a time interval \([r, t]\), the average velocity \(\mathbf{u}\) is defined as:
\begin{equation}
\mathbf{u}(\mathbf{z}_t, r, t) = \frac{1}{t - r} \int_{r}^{t} \mathbf{v}(\mathbf{z}_t | \mathbf{x}, \boldsymbol{\epsilon}) \, \mathrm{d}\tau.
\end{equation}
This representation allows for single-step sampling from \(\mathbf{z}_t\) to \(\mathbf{z}_r\), since \(\mathbf{z}_r = \mathbf{z}_t - (t - r) \mathbf{u}(\mathbf{z}_t, r, t)\).

To train a network \(\mathbf{u}_\theta\) that predicts this quantity, we derive a self-supervised objective. Differentiating the definition of \(\mathbf{u}\) with respect to the end time \(t\) yields the MeanFlow Identity:
\begin{equation}
\mathbf{u}(\mathbf{z}_t, r, t) = \mathbf{v}(\mathbf{z}_t | \mathbf{x}, \boldsymbol{\epsilon}) - (t - r) \frac{\mathrm{d}}{\mathrm{d}t} \mathbf{u}(\mathbf{z}_t, r, t),
\end{equation}
where the total derivative expands as \(\frac{\mathrm{d}}{\mathrm{d}t}\mathbf{u} = \partial_t \mathbf{u} + \mathbf{v}_t(\mathbf{z}_t | \mathbf{x}, \boldsymbol{\epsilon})\cdot \partial_{\mathbf{z}_t} \mathbf{u}\). This identity provides a target for the average velocity, leading to the MeanFlow objective:
\begin{equation}\label{eq:meanflow-loss}
\mathcal{L}_{\text{MF}} = \| \mathbf{u}_\theta(\mathbf{z}_t, r, t) - \text{sg}(\mathbf{u}_{\text{tgt}}) \|^2,
\end{equation}
with the target defined as
\begin{equation}
\mathbf{u}_{\text{tgt}} = \mathbf{v}(\mathbf{z}_t | \mathbf{x}, \boldsymbol{\epsilon}) - (t - r) \left[ \partial_t \mathbf{u}_\theta + \mathbf{v}(\mathbf{z}_t | \mathbf{x}, \boldsymbol{\epsilon}) \cdot \partial_{\mathbf{z}_t} \mathbf{u}_\theta \right].
\end{equation}
Here, \(\text{sg}(\cdot)\) denotes the stop-gradient operation. This formulation enables \(\mathbf{u}_\theta\) to be trained in a self-supervising manner: the network learns to predict the average velocity that must satisfy the kinematic identity dictated by the underlying instantaneous velocity field \(\mathbf{v}\).

\subsubsection{Improved MeanFlow} 
MeanFlow objective, defined in Equation \ref{eq:meanflow-loss}, relies on a target $\mathbf{u}_{\text{tgt}}$ that depends recursively on the network's own predictions via the stop-gradient operation. This self-supervising structure on the average velocity can, in practice, lead to training instability and suboptimal gradient dynamics~\cite{geng2025improved}.

To address this, the i-MF objective was introduced, which reformulates the learning problem into a standard regression loss on the instantaneous velocity~\cite{geng2025improved}. Starting from the MeanFlow identity, we rearrange terms to define a new network-predicted quantity:
\begin{equation}
\mathbf{V}_\theta(\mathbf{z}_t, r, t) \equiv \mathbf{u}_\theta + (t - r) \ \text{sg}\left(\left[\, \partial_t \mathbf{u}_\theta + \mathbf{v}(\mathbf{z}_t | \mathbf{x}, \boldsymbol{\epsilon})  \partial_{\mathbf{z}_t} \mathbf{u}_\theta \,\right]\right).
\end{equation}
Substituting this definition back into the MeanFlow identity shows that a perfect model should satisfy $\mathbf{V}_\theta(\mathbf{z}_t, r, t) = \mathbf{v}(\mathbf{z}_t | x, \epsilon)$. This yields a direct and stable regression objective:
\begin{equation}\label{eq:i-mf}
\mathcal{L}_{\text{i-MF}} = \mathbb{E}_{t, \mathbf{x}, \boldsymbol{\epsilon}} \| \mathbf{V}_\theta(\mathbf{z}_t, r, t) - \mathbf{v}(\mathbf{z}_t | \mathbf{x}, \boldsymbol{\epsilon}) \|^2.
\end{equation}
This formulation moves the self-supervising signal to instantaneous velocity, resulting in a conventional conditional flow matching loss. Consequently, the i-MF objective improves both training stability and final sampling quality compared to the original MeanFlow approach on generative tasks~\cite{geng2025improved}.

\subsection{General Algorithm}
The proposed framework operates in two sequential training stages to balance computational efficiency with high-fidelity generative refinement. In the first stage, a neural operator backbone is trained to learn the core temporal dynamics of the system in a low-resolution latent, as shown in Figure \ref{fig:MENO_framework} (b). This model, trained via a standard autoregressive objective (e.g., mean squared error on future states), provides accurate but coarse-grained future predictions. The second stage focuses on enhancing spatial resolution. A separate MeanFlow-Enhanced Neural Operator decoder is trained to perform a one-step mapping from these corrupted latent states to high-resolution physical fields, demonstrated in Figure \ref{fig:MENO_framework} (a) and Algorithm \ref{alg:training}, which does not explicitly condition on low-resolution latent states. This flexible modular design enables the framework to be adapted easily to any existing NO pipeline. This decoder is optimized using the $\mathcal{L}_\text{i-MF}$ loss (Equation \ref{eq:i-mf}). During inference, the pre-trained NO autoregressively rolls out a low-resolution trajectory, and the MENO decoder acts as a one-step generative refining module on each predicted frame, yielding a high-resolution, physically consistent forecast. This is summarized in Figure \ref{fig:MENO_framework} (c) and Algorithm \ref{alg:inference}. 

\begin{algorithm}[h]
\caption{MENO Decoder Training}
\label{alg:training}
\begin{algorithmic}[1]
\REQUIRE high-resolution dataset $\mathcal{D}$ (empirical distribution $\hat p_{\mathcal D}$), decoder $\mathbf{u}_\theta$, iterations $K$, batch size $B$, learning rate $\eta$
\STATE Initialize $\theta$
\FOR{$k=1,\dots,K$}
    \STATE Sample $\{\mathbf{x}^{(i)}\}_{i=1}^B \sim \hat p_{\mathcal D}$, $\boldsymbol{\epsilon}^{(i)}\sim\mathcal N(0,I)$, $t^{(i)}$, $r^{(i)}$
    \STATE $\mathbf{z}_{t}^{(i)}\gets (1-t^{(i)})\mathbf{x}^{(i)}+t^{(i)}\boldsymbol{\epsilon}^{(i)}$, \quad $\mathbf{v}^{(i)}\gets \boldsymbol{\epsilon}^{(i)}-\mathbf{x}^{(i)}$
    \STATE $\mathbf{V}_\theta^{(i)}\gets \mathbf{u}_\theta(\mathbf{z}_t^{(i)},r^{(i)},t^{(i)})+(t^{(i)}-r^{(i)})\cdot\text{sg}\Big(\partial_t \mathbf{u}_\theta+{\mathbf{v}^{(i)}}^{\!\top}\nabla_{\mathbf{z}}\mathbf{u}_\theta\Big)$
    \STATE $\mathcal{L}_{\text{i-MF}}\gets \frac{1}{B}\sum_{i=1}^B \|\mathbf{V}_\theta^{(i)}-\mathbf{v}^{(i)}\|_2^2$
    \STATE $\theta \gets \theta-\eta\nabla_\theta\big(\mathcal{L}_{\text{i-MF}}\big)$
\ENDFOR
\end{algorithmic}
\end{algorithm}

\begin{algorithm}[t]
\caption{MENO Inference}
\label{alg:inference}
\begin{algorithmic}[1]
\REQUIRE low-resolution initial state $\mathbf{a}_0^{\text{LR}}$, trained models $\mathcal{G}_{\phi^*}$, $\mathbf{u}_{\theta^*}$, horizon $T$, uniform upsampler $U(\cdot)$, noise strength $\tau$
\FOR{$t=1,\dots,T$}
\STATE $\tilde{\mathbf{a}}^\text{LR}_t \leftarrow \mathcal{G}_{\phi^*}(\tilde{\mathbf{a}}^\text{LR}_{t-1})$
\STATE Sample $\boldsymbol{\epsilon}_t \sim \mathcal{N}(0,I)$, $\mathbf{z}_t = (1-\tau)U(\tilde{\mathbf{a}}^\text{LR}_t)+ \tau \boldsymbol{\epsilon}_t$
\STATE $\hat{\mathbf{x}}_t \leftarrow \mathbf{z}_t - \mathbf{u}_{\theta^*}(\mathbf{z}_t, 0, \tau) \tau$
\ENDFOR
\end{algorithmic}
\textbf{Return:} High-resolution forecast $\{\hat{\mathbf{x}}_t\}_{t=1}^T$.
\end{algorithm}

\subsection{Evaluation metrics}
\label{subsec:metrics}

We evaluate predictions through autoregressive time-series rollouts initialized
from the low-resolution ground-truth initial condition, and compare against the
corresponding high-resolution ground-truth trajectories using three metrics.

The relative $L_2$ error measures pointwise reconstruction accuracy:
\begin{equation}
\mathrm{R}L_{2}(k)
=
\frac{1}{N}
\sum_{j=1}^{N}
\frac{
\left\|
\hat{z}^{\,j}_{k}-z^{\,j}_{k}
\right\|_2
}{
\left\|
z^{\,j}_{k}
\right\|_2
},
\label{eq:rl2}
\end{equation}
where $\hat{z}^{\,j}_{k}$ and $z^{\,j}_{k}$ are the predicted and ground-truth
states at time step $k$ for trajectory $j$.

The Structural Similarity Index Measure (SSIM) evaluates local structural
agreement:
\begin{equation}
\mathrm{SSIM}(x,y)
=
\frac{
(2\mu_x\mu_y+C_1)(2\sigma_{xy}+C_2)
}{
(\mu_x^2+\mu_y^2+C_1)(\sigma_x^2+\sigma_y^2+C_2)
}.
\label{eq:ssim}
\end{equation}
It is computed within a Gaussian window and averaged over the spatial domain.

Finally, the Power Spectrum Density Discrepancy (PSDD) measures agreement of
the spatial-frequency distribution:
\begin{equation}
\mathcal{L}_{\mathrm{PSD}}(x,y)
=
\frac{1}{B}
\sum_{b=1}^{B}
\frac{1}{HWC}
\sum_{i,j,c}
\left|
\widehat{P}^{(b)}_{x,i,j,c}
-
\widehat{P}^{(b)}_{y,i,j,c}
\right|,
\label{eq:psdd}
\end{equation}
where $\widehat{P}=P/\sum P$ is the normalized two-dimensional Fourier power
spectrum. Relative $L_2$ and SSIM are reported over short prediction horizons to
emphasize instantaneous accuracy, whereas PSDD is computed over full rollouts to
assess long-time statistical fidelity.

\section{Experiments}
\label{section:experiments}

We evaluate MENO on three PDE-governed dynamical systems that span phase
separation (Cahn-Hilliard), turbulent transport (2D Kolmogorov flow), and
non-equilibrium active fluids (active matter). For each dataset we consider two
cross-resolution settings, written as
low-resolution~$\rightarrow$~high-resolution; for example,
$20\rightarrow100$ indicates that the neural operator is trained on
$20\times 20$ fields and evaluated on the $100\times 100$ grid. Baseline neural
operators are trained on low-resolution data and evaluated autoregressively;
high-resolution predictions are obtained either by direct neural operator
super-resolution or by applying a generative decoder (a diffusion model or the
proposed MeanFlow decoder) to the low-resolution rollout. The generative
decoders are trained once per dataset on high-resolution fields and reused
across different low-resolution rollout settings, since the decoder operates on
the high-resolution grid.

\subsection{Baselines}
\label{subsec:baselines}

We compare MENO against direct neural operator super-resolution models and
diffusion-enhanced neural operators. The neural operator backbones are FNO and
UNO. UNO can be viewed as a multi-scale extension of FNO that incorporates
skip connections and a U-Net-inspired architecture, often improving
reconstruction fidelity for spatially structured fields.

As a generative refinement baseline, we use a diffusion-model decoder trained
with the DDPM objective. The forward noising process is given by
\begin{equation}
\mathbf{x}_t =
\sqrt{\bar{\alpha}_t}\,\mathbf{x}_0
+
\sqrt{1-\bar{\alpha}_t}\,\boldsymbol{\epsilon},
\qquad
\boldsymbol{\epsilon}\sim\mathcal{N}(0,I),
\label{eq:ddpm_forward}
\end{equation}
where $\bar{\alpha}_t=\prod_{s=1}^{t}(1-\beta_s)$ and
$\{\beta_t\}_{t=1}^{T}$ is the noise schedule. At inference, sampling is
accelerated using DDIM, a deterministic non-Markovian reverse trajectory with
skipped denoising steps (Appendix~\ref{appendix:dmno}). All models of the same
type share identical hyperparameter settings; full configurations and
sensitivity studies for the MENO noise level and the DDIM step count are
reported in
Appendices~\ref{appendix:hyperparameters} and~\ref{appendix:denoise}.

\begin{figure*}[h]
    \centering
    \includegraphics[width=0.96\linewidth]{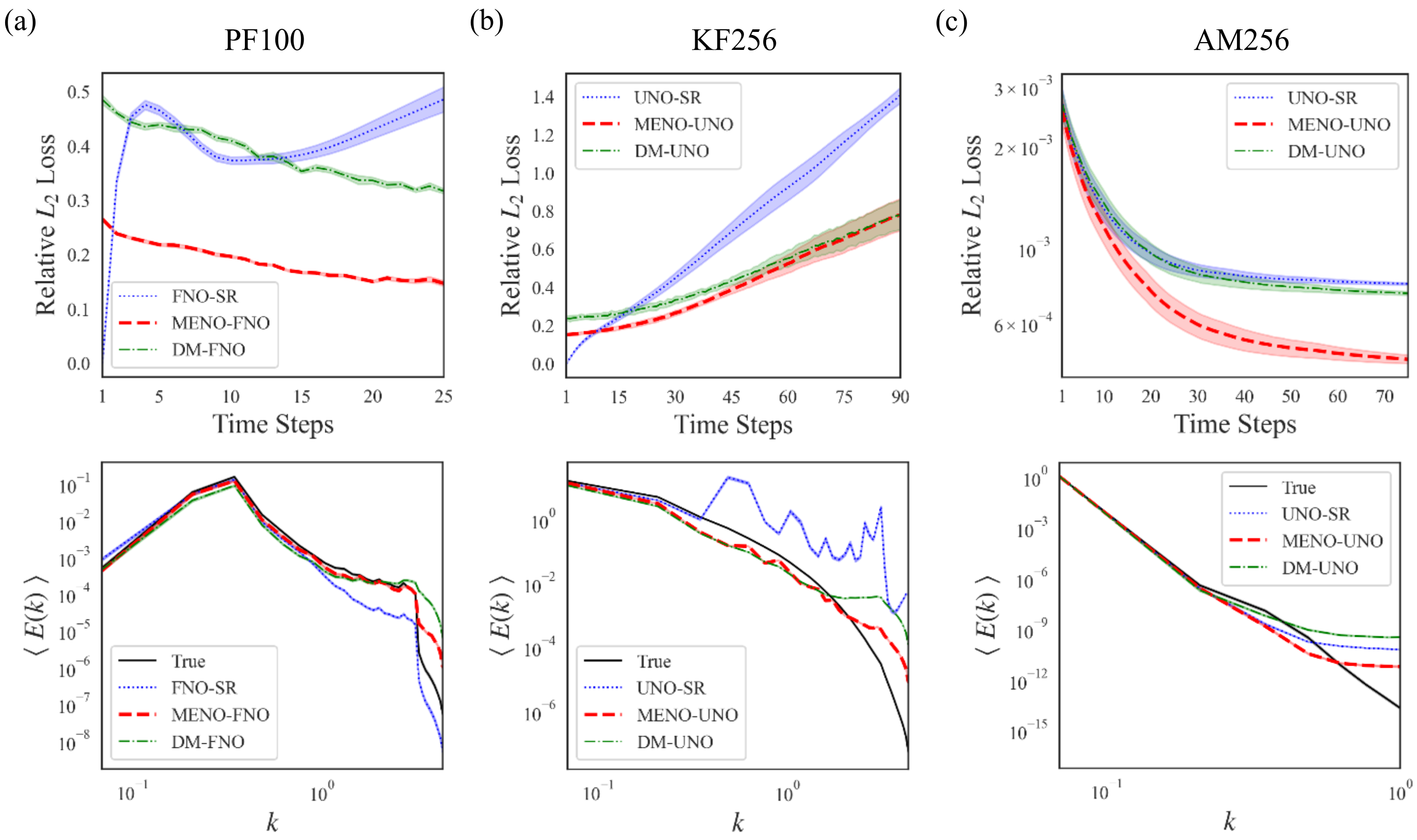}
    \caption{Relative \(L_2\) loss over rollout time (top row) and the
    corresponding wavenumber energy spectra (bottom row). Shaded regions
    denote the standard error of the mean (SEM) computed over multiple test
    trajectories (PF100: 10, KF256: 8, AM256: 25). (a) PF100
    \(20\rightarrow 100\): a FNO model trained on \(20\times 20\) data and its
    enhanced variants. (b) KF256 \(64\rightarrow 256\): performance of a UNO
    model trained on \(64\times 64\) data and its enhanced variants. (c) AM256
    \(64\rightarrow 256\): performance of a UNO model trained on
    \(64\times 64\) data and its enhanced variants.}
    \label{fig:l2_and_spec}
    \vspace{-1em}
\end{figure*}

\subsection{Cahn-Hilliard phase field}
\label{Exp:pf}

\begin{table}[t]
\centering
\caption{PF100 prediction metrics for direct neural operator super-resolution,
diffusion-enhanced neural operators, and MENO. The best value in each column is
shown in bold. The stand-alone UNO super-resolution model was not included for
PF100 because it did not produce reliable autoregressive rollouts under the same
training setting as FNO. Relative improvement is computed with respect to the
corresponding direct neural operator baseline.}
\label{tab:pf100_metrics}
\footnotesize
\setlength{\tabcolsep}{4pt}
\begin{tabular}{lcccccc}
\toprule
& \multicolumn{3}{c}{$20\rightarrow100$}
& \multicolumn{3}{c}{$50\rightarrow100$} \\
\cmidrule(lr){2-4} \cmidrule(lr){5-7}
Model
& $\mathrm{R}L_2 \downarrow$
& SSIM $\uparrow$
& PSDD $\downarrow$
& $\mathrm{R}L_2 \downarrow$
& SSIM $\uparrow$
& PSDD $\downarrow$ \\
\midrule
\multicolumn{7}{l}{\textit{Neural operators without enhancement}} \\
FNO
& 0.64
& 0.60
& $4.98\times10^{-5}$
& 0.26
& 0.85
& $1.99\times10^{-5}$ \\
\midrule
\multicolumn{7}{l}{\textit{Diffusion-enhanced neural operators}} \\
DM-FNO
& 0.50
& 0.65
& $2.97\times10^{-5}$
& 0.30
& 0.77
& $2.28\times10^{-5}$ \\
DM-UNO
& 0.54
& 0.63
& $3.34\times10^{-5}$
& 0.32
& 0.76
& $2.40\times10^{-5}$ \\
\midrule
\multicolumn{7}{l}{\textit{MENO models}} \\
MENO-FNO
& \textbf{0.21}
& \textbf{0.86}
& $\mathbf{1.59\times10^{-5}}$
& \textbf{0.08}
& \textbf{0.96}
& $\mathbf{6.52\times10^{-6}}$ \\
MENO-UNO
& 0.29
& 0.82
& $2.15\times10^{-5}$
& 0.13
& 0.94
& $9.89\times10^{-6}$ \\
\midrule
Relative improvement (\%)
& 67.2
& 43.3
& 68.1
& 69.2
& 12.9
& 67.2 \\
\bottomrule
\end{tabular}
\end{table}

We first consider the Cahn-Hilliard phase-field system, a standard benchmark
for coarsening dynamics and phase separation~\cite{xue2025equivariant}. It
stresses surrogate models in two complementary ways: the order parameter
develops sharp diffuse interfaces whose width is set by $\lambda$, and the
system is dissipative, so any reasonable model must reproduce the monotonic
decay of a global thermodynamic free energy. The model describes the evolution
of an order parameter $\phi(\mathbf{x},t)$ for a binary mixture:
\begin{equation}
\frac{\partial \phi}{\partial t}
=
\nabla \cdot \left(M\nabla \mu\right),
\label{eq:ch}
\end{equation}
where $M$ is the mobility and $\mu$ is the chemical potential. The chemical
potential is derived from the Ginzburg-Landau free energy
\begin{equation}
\mathcal{F}[\phi]
=
\int_{\Omega}
\left(
\frac{\lambda}{\varepsilon}W(\phi)
+
\frac{\lambda\varepsilon}{2}
|\nabla\phi|^2
\right)
\,\mathrm{d}\mathbf{x},
\qquad
W(\phi)=\frac{1}{4}(\phi^2-1)^2,
\label{eq:free-energy}
\end{equation}
which gives the chemical potential
\begin{equation}
\mu
=
\lambda
\left(
\frac{\phi(\phi^2-1)}{\varepsilon}
-
\varepsilon\nabla^2\phi
\right).
\label{eq:ch_mu}
\end{equation}
We generate 300 independent simulations on the unit square
$\Omega=[0,1]^2$ at $100\times100$ resolution using COMSOL Multiphysics, with
constant $M=1$ and $\lambda=0.01$. The resolution-transfer settings are
$20\rightarrow100$ and $50\rightarrow100$.

\begin{figure}[!h]
    \centering
    \includegraphics[width=0.85\linewidth]{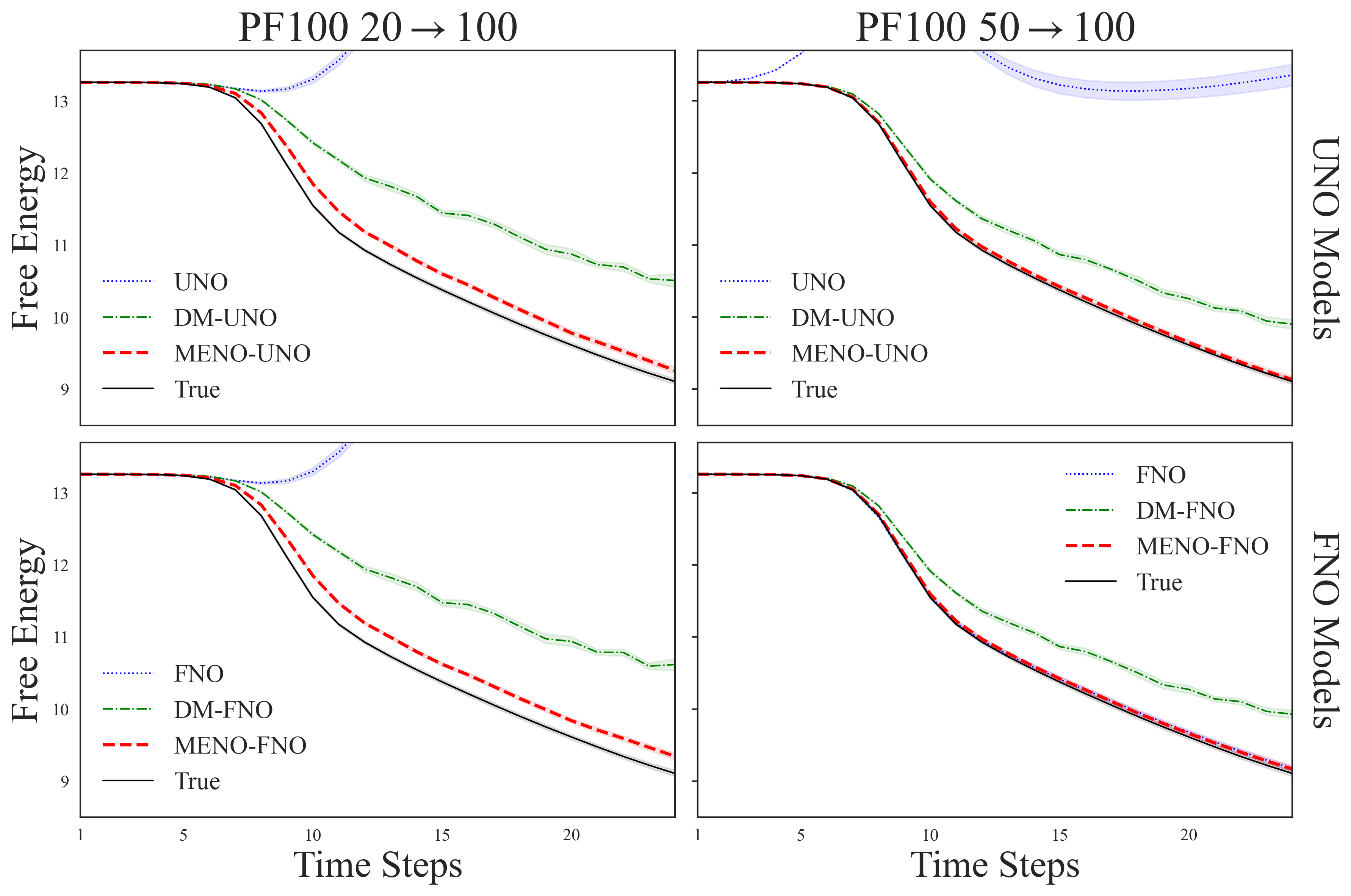}
    \caption{Free-energy trajectories for all models and resolution settings
    on PF100. Upper panels: UNO-based models; lower panels: FNO-based models.
    Left column: $20 \rightarrow 100$; right column: $50 \rightarrow 100$.
    Only the MENO variants consistently recover the correct free-energy
    evolution. Shaded regions denote the SEM over 10 test trajectories.}
    \label{fig:free_energy}
\end{figure}

\begin{figure}[!h]
    \centering
    \includegraphics[width=0.85\linewidth]{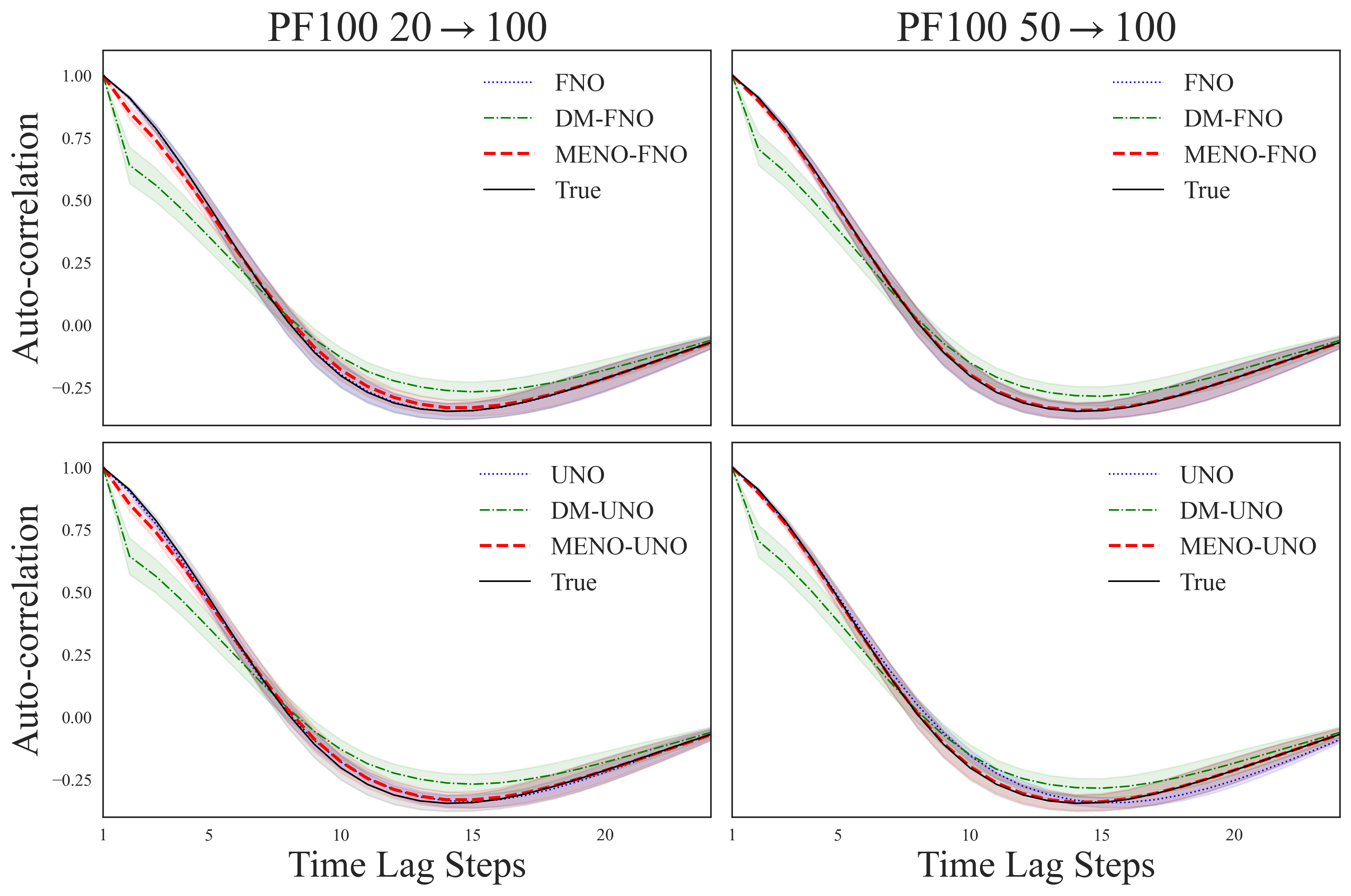}
    \caption{Temporal autocorrelation functions on PF100 across all resolution
    settings and model variants. Shaded bands denote the SEM over 10 test
    trajectories. MENO best preserves the long-time temporal coherence of the
    phase-field dynamics.}
    \label{fig:pf100_autocorr}
\end{figure}

Table~\ref{tab:pf100_metrics} reports the PF100 prediction metrics for direct
neural operator super-resolution, diffusion-enhanced neural operators, and
MENO. The most severe setting is $20\rightarrow100$, where the low-resolution
input contains only a coarse representation of the diffuse-interface geometry.
In this regime, direct FNO super-resolution produces a relative $L_2$ error of
0.64 and an SSIM of 0.60, indicating that the rollout captures the large-scale
phase morphology but loses substantial interfacial detail. Diffusion refinement
improves the FNO baseline, reducing the relative $L_2$ error to 0.50 and the
PSDD to $2.97\times10^{-5}$, but it does not fully recover the missing mid- and
high-wavenumber content. MENO-FNO gives the strongest improvement, reducing the
relative $L_2$ error to 0.21, increasing SSIM to 0.86, and lowering PSDD to
$1.59\times10^{-5}$. These gains show that the proposed single-step decoder
improves both pointwise agreement and structural fidelity, rather than merely
sharpening the visual appearance of the fields.

\begin{figure}[t]
    \centering
    \includegraphics[width=0.95\linewidth]{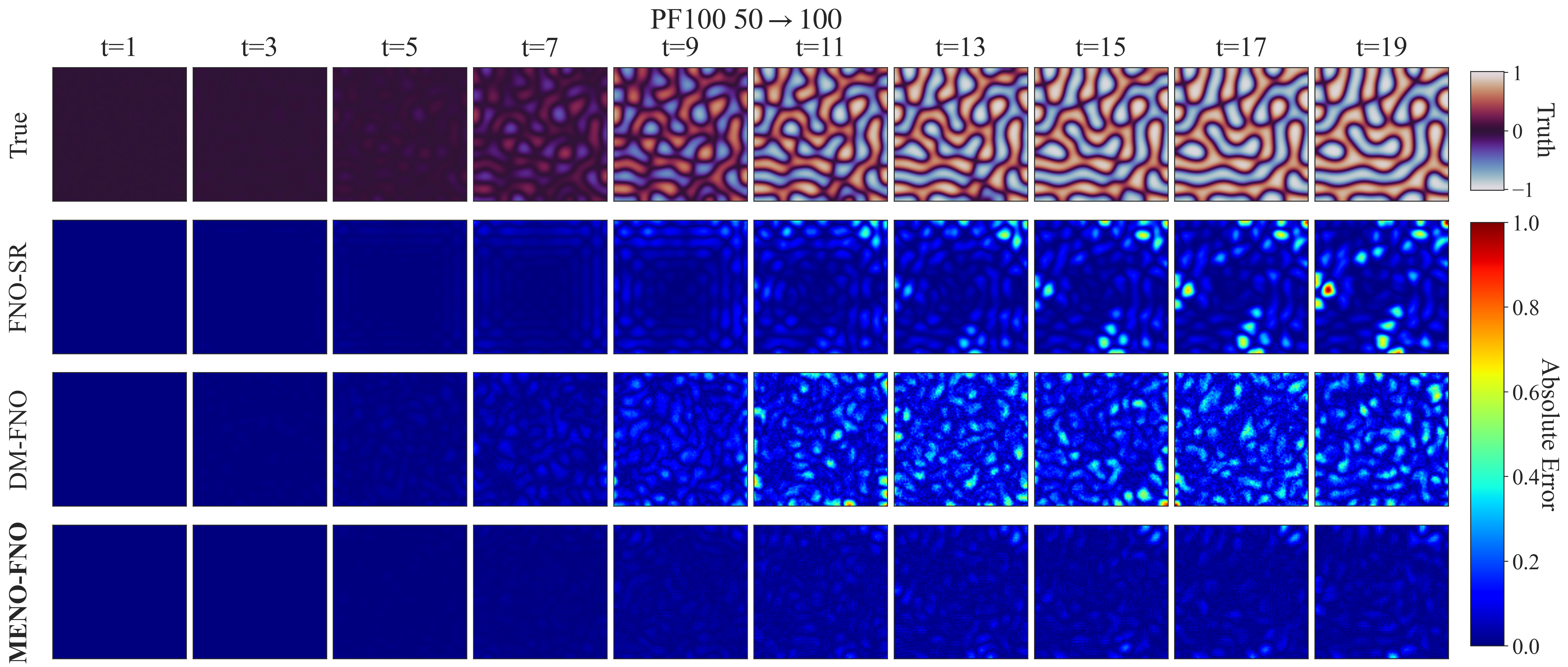}
    \caption{Snapshots along an uncurated PF100 trajectory for the
    $50\rightarrow100$ setting. The top row presents the high-resolution ground
    truth, and the next three rows visualize absolute errors for NO-SR,
    MENO-NO, and DM-NO, respectively. MENO produces the lowest interface-localized
    error and best preserves the evolving phase morphology.}
    \label{fig:pf100_vis}
\end{figure}

The same trend becomes more pronounced in the less under-resolved
$50\rightarrow100$ setting. Because the coarse input already contains more
information about the interface locations, the main challenge is to reconstruct
the correct interfacial width and small-scale curvature without introducing
spurious oscillations. MENO-FNO reduces the relative $L_2$ error from 0.26 to
0.08 and improves SSIM from 0.85 to 0.96, while also reducing PSDD from
$1.99\times10^{-5}$ to $6.52\times10^{-6}$. The diffusion-enhanced baselines,
by contrast, do not consistently improve the direct FNO prediction in this
setting: both DM-FNO and DM-UNO have larger relative $L_2$ errors and lower
SSIM than the direct FNO baseline. This suggests that iterative denoising can
introduce visually plausible but dynamically inconsistent corrections when the
low-resolution rollout is already moderately accurate.

Figure~\ref{fig:l2_and_spec}(a) supports this interpretation. Over rollout
time, MENO maintains the lowest relative $L_2$ error and avoids the error growth
observed in the direct super-resolution baseline. In spectral space, MENO
follows the ground-truth energy distribution over a broader wavenumber range,
whereas direct FNO-SR loses energy in the smaller scales and the
diffusion-enhanced model only partially restores it. For Cahn-Hilliard
dynamics, this spectral agreement is important because the evolution of domain
boundaries depends on accurately resolving interface curvature and local
chemical-potential gradients.

Qualitative rollout comparisons for the $50\rightarrow100$ setting are shown
in Figure~\ref{fig:pf100_vis}. Direct neural operator super-resolution captures
the coarse phase domains but produces larger errors near interfaces, where
small shifts in boundary position or thickness strongly affect the field-level
metrics. Diffusion-enhanced refinement reduces some of these localized errors
but does not consistently align the reconstructed interfaces with the
ground-truth morphology. MENO yields lower error magnitude along the evolving
phase boundaries, indicating that it better recovers the geometry and sharpness
of the diffuse interfaces. This qualitative behavior is consistent with the
large SSIM and PSDD improvements in Table~\ref{tab:pf100_metrics}.

The Cahn-Hilliard equation is a dissipative gradient-flow system, and its total
free energy should decrease monotonically over time. We therefore compute the
discretized Ginzburg-Landau free energy for each predicted trajectory and
compare it with the ground truth. As shown in Figure~\ref{fig:free_energy},
direct neural operator super-resolution rollouts drift away from the correct
dissipative trend at long horizons, indicating that the predicted morphology is
not thermodynamically consistent even when the fields remain visually
plausible. Diffusion-enhanced refinement reduces part of this drift but still
fails to match the ground-truth energy trajectory. In contrast, MENO closely
tracks the ground-truth free-energy decay across both resolution settings and
both neural operator backbones, showing that the recovered interfaces are not
only sharper but also more consistent with the dissipative structure of the
governing equation.

To further assess whether the learned surrogate preserves long-time temporal
statistics, we compute the temporal autocorrelation function of the predicted
phase fields. The autocorrelation is evaluated along the time axis using a
Fourier-transform implementation with de-meaning and zero-padding to avoid
wrap-around effects, as detailed in Appendix~\ref{appendix:metrics}.
Figure~\ref{fig:pf100_autocorr} shows that direct super-resolution models
decorrelate incorrectly as the rollout proceeds, reflecting a mismatch in the
rate at which phase morphology evolves during coarsening. Diffusion-enhanced
refinement partially improves the decay profile but still deviates from the
ground truth at longer lags. MENO remains closest to the ground-truth
autocorrelation over the full lag range, indicating that it preserves the time
scale of domain evolution as well as the instantaneous spatial structure.
Taken together, the PF100 results show that MENO addresses the central
difficulty of the Cahn-Hilliard benchmark: it reconstructs under-resolved
diffuse interfaces while preserving the thermodynamic and temporal signatures
of coarsening dynamics.

\begin{table}[t]
\centering
\caption{PF100 model size and inference time for generative-enhanced neural
operators. Timings are measured per predicted frame for the full pipeline,
including the neural operator and the decoder.}
\label{tab:pf100_size_time}
\footnotesize
\setlength{\tabcolsep}{8pt}
\begin{tabular}{lcc}
\toprule
Model
& Parameters, NO + decoder
& Inference time per frame (s) \\
\midrule
\multicolumn{3}{l}{\textit{Diffusion-enhanced neural operators}} \\
DM-FNO
& 6.33M + 1.44M
& $0.176 \pm 0.022$ \\
DM-UNO
& 6.38M + 1.44M
& $0.178 \pm 0.021$ \\
\midrule
\multicolumn{3}{l}{\textit{MENO models}} \\
MENO-FNO
& 6.33M + 1.44M
& $\mathbf{0.028 \pm 0.022}$ \\
MENO-UNO
& 6.38M + 1.44M
& $0.031 \pm 0.021$ \\
\bottomrule
\end{tabular}
\end{table}

Finally, Table~\ref{tab:pf100_size_time} shows that the accuracy and physics
improvements above do not come at the cost of a larger model. The MENO and
diffusion-enhanced variants use matched neural operator and decoder parameter
budgets: for example, DM-FNO and MENO-FNO both use a 6.33M-parameter FNO and a
1.44M-parameter decoder, while DM-UNO and MENO-UNO use comparable
6.38M-parameter UNO backbones with the same decoder size. The difference is
therefore in the inference procedure rather than in model capacity. Because
MENO performs single-step generative refinement instead of iterative DDIM
sampling, MENO-FNO reduces the per-frame inference time from
$0.176\pm0.022$~s to $0.028\pm0.022$~s, corresponding to a $6.3\times$
speedup. MENO-UNO similarly reduces the time from $0.178\pm0.021$~s to
$0.031\pm0.021$~s, a $5.7\times$ speedup. Thus, on PF100, MENO improves
interface reconstruction, free-energy consistency, and temporal statistics
while also substantially reducing inference cost at fixed parameter budget.

\subsection{Kolmogorov flow}
\label{Exp:kf}

We next evaluate MENO on two-dimensional Kolmogorov
flow~\cite{chandler2013invariant}, a canonical benchmark for turbulent,
statistically stationary
dynamics~\cite{pope2001turbulent,hairer2006ergodicity,Kochkov2021-ML-CFD}.
Unlike Cahn-Hilliard, it has a broadband energy spectrum extending into the
small scales and does not relax to a steady state, so the surrogate must
reproduce the high-wavenumber tail of the spectrum at every step rather than at
a single coarsening end-state. The system is governed by the incompressible
Navier-Stokes equations with spatially periodic forcing:
\begin{equation}
\frac{\partial \mathbf{u}}{\partial t}
+
\mathbf{u}\cdot\nabla\mathbf{u}
=
-\nabla p
+
\nu\nabla^2\mathbf{u}
+
\mathbf{f}(\mathbf{x}),
\qquad
\nabla\cdot\mathbf{u}=0,
\label{eq:kolmogorov_ns}
\end{equation}
with forcing $\mathbf{f}(\mathbf{x})=(\sin(ny),0)$.
The flow is evolved on $\Omega=[0,2\pi]^2$ at Reynolds number
$\mathrm{Re}=1000$ and $256\times256$ resolution. The dataset contains 70
trajectories, each with 180 vorticity-field frames. We consider the
$32\rightarrow256$ and $64\rightarrow256$ settings.

Table~\ref{tab:kf256_metrics} summarizes the KF256 results. Compared with
PF100, this benchmark is more demanding spectrally because the vorticity field
contains broadband turbulent structure and must maintain a statistically
stationary cascade rather than relax toward a smooth final state. In the
strongly under-resolved $32\rightarrow256$ setting, direct FNO achieves the
lowest relative $L_2$ error, 0.19, while MENO-FNO and MENO-UNO obtain slightly
larger values of 0.21. However, this pointwise metric alone does not capture
the full behavior of the rollout. MENO improves SSIM from 0.73 to 0.76 and
reduces PSDD from $1.22\times10^{-5}$ to $8.62\times10^{-6}$ relative to the
FNO baseline, indicating better preservation of coherent vortical structure
and spectral content. The $32\rightarrow256$ case therefore exposes a tradeoff:
a smoother direct prediction can achieve a slightly lower averaged pixel-wise
error, while MENO better matches the spatial statistics of the turbulent field.

In the less extreme $64\rightarrow256$ setting, where the coarse rollout
contains more reliable information about the large-scale vorticity dynamics,
MENO improves all metrics simultaneously. MENO-UNO reduces the relative $L_2$
error to 0.08, compared with 0.13 for FNO and 0.22 for UNO, while increasing
SSIM to 0.93 and reducing PSDD to $5.16\times10^{-6}$. The improvement in PSDD
is particularly important for Kolmogorov flow because errors at high
wavenumbers correspond to missing or distorted small-scale vortical structures.
The diffusion-enhanced baselines improve the spectral metric relative to direct
neural operators in some cases, but they do so with substantially worse
relative $L_2$ and SSIM, suggesting that the iterative decoder can inject
high-frequency content that is not consistently aligned with the true vorticity
field.

\begin{figure*}[t]
    \centering
    \includegraphics[width=0.95\linewidth]{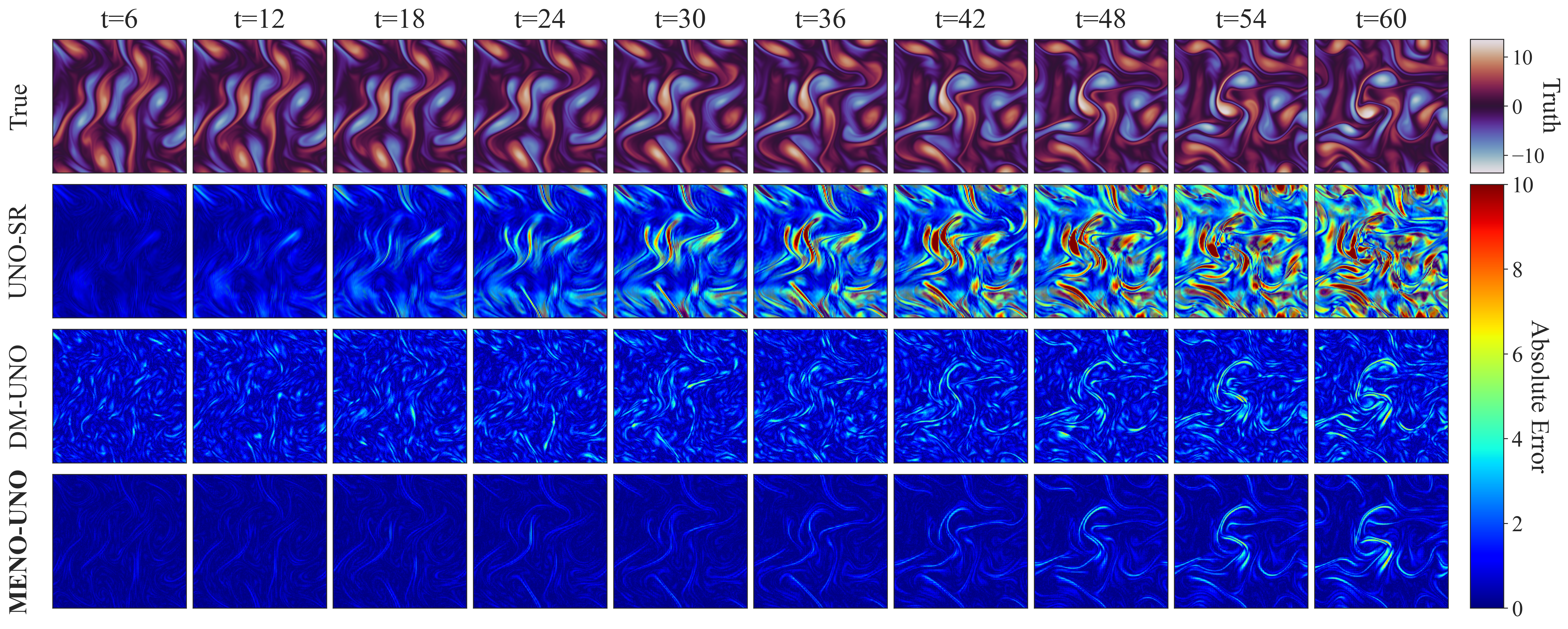}
    \caption{Snapshots along a trajectory from the KF256 dataset at
    $t=6,12,\ldots,54$. The top row reports the high-resolution ground truth at
    $256\times256$. The next three rows visualize the absolute errors of
    UNO-SR, MENO-UNO, and DM-UNO, respectively. The color bars indicate the
    vorticity field and absolute error magnitude.}
    \label{fig:kf_vis}
\end{figure*}

The qualitative rollout in Figure~\ref{fig:kf_vis} illustrates these
differences directly. UNO-SR accumulates error around regions of strong
vorticity gradients and increasingly misses fine vortical filaments as the
trajectory advances. DM-UNO reduces some high-frequency discrepancy but leaves
larger localized error patches, consistent with a refinement process that adds
texture without fully correcting the underlying low-resolution dynamics.
MENO-UNO maintains smaller error magnitudes over the rollout and better
preserves the shape and placement of coherent vortices. This behavior is
important because turbulent prediction errors are rapidly amplified when
small-scale structures are misplaced, even if the large-scale flow remains
qualitatively recognizable.

Figure~\ref{fig:l2_and_spec}(b) further clarifies the role of the generative
decoder. In the time-resolved error curves, MENO suppresses the late-time
growth observed in the direct neural operator baseline. In the corresponding
energy spectra, MENO gives the closest match to the ground-truth
high-wavenumber tail, whereas direct super-resolution loses energy at small
scales and diffusion-enhanced refinement does not consistently align the
recovered spectral content with the true field. Thus, the main advantage of
MENO on KF256 is not simply lower frame-wise error; it is the ability to
recover small-scale turbulent content while keeping it dynamically synchronized
with the low-resolution rollout.

We also evaluate temporal autocorrelation to test whether the models reproduce
the statistically stationary temporal structure of the turbulent flow. The
autocorrelation is computed from the predicted vorticity fields along the time
axis using the Fourier-transform procedure described in
Appendix~\ref{appendix:metrics}. Figure~\ref{fig:kf256_autocorr} shows that
direct neural operator super-resolution decorrelates too quickly in several
settings, which indicates that the predicted vortical structures lose temporal
coherence faster than in the true flow. Diffusion-enhanced models partially
correct this behavior but still deviate from the ground-truth decay. MENO gives
the closest autocorrelation profile, especially at early and intermediate lags
where coherent vortex evolution dominates the memory of the flow. The KF256
results therefore support the motivation for using a generative refinement
step: MENO reconstructs missing turbulent scales without sacrificing temporal
coherence or requiring the expensive multi-step sampling used by diffusion
models.

\begin{figure}[!h]
    \centering
    \includegraphics[width=0.85\linewidth]{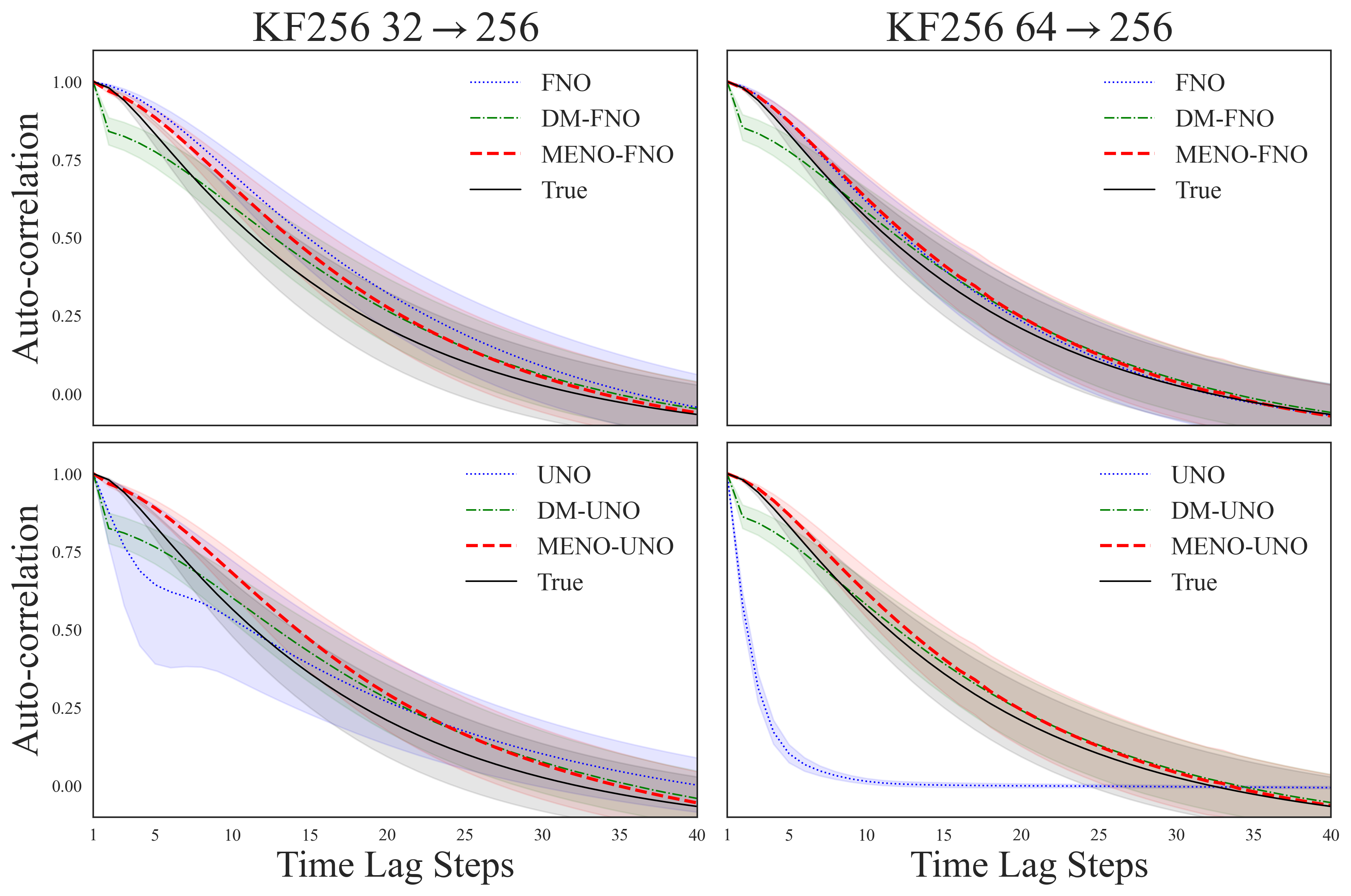}
    \caption{Temporal autocorrelation functions on KF256 across all resolution
    settings and model variants. Shaded bands denote the SEM over 8 test
    trajectories. The early-time decay is emphasized because it captures the
    dominant temporal memory of the turbulent vorticity field.}
    \label{fig:kf256_autocorr}
\end{figure}

\begin{table}[t]
\centering
\caption{KF256 prediction metrics for direct neural operator super-resolution,
diffusion-enhanced neural operators, and MENO. $\mathrm{R}L_2$ and SSIM are
computed over the first 20 frames, while PSDD is computed over the full
180-frame trajectories. The best value in each column is shown in bold.
Relative improvement is computed with respect to the corresponding direct
neural operator baseline.}
\label{tab:kf256_metrics}
\footnotesize
\setlength{\tabcolsep}{4pt}
\begin{tabular}{lcccccc}
\toprule
& \multicolumn{3}{c}{$32\rightarrow256$}
& \multicolumn{3}{c}{$64\rightarrow256$} \\
\cmidrule(lr){2-4} \cmidrule(lr){5-7}
Model
& $\mathrm{R}L_2 \downarrow$
& SSIM $\uparrow$
& PSDD $\downarrow$
& $\mathrm{R}L_2 \downarrow$
& SSIM $\uparrow$
& PSDD $\downarrow$ \\
\midrule
\multicolumn{7}{l}{\textit{Neural operators without enhancement}} \\
FNO
& $\mathbf{0.19}$
& 0.73
& $1.22\times10^{-5}$
& 0.13
& 0.82
& $1.00\times10^{-5}$ \\
UNO
& 0.23
& 0.66
& $1.74\times10^{-5}$
& 0.22
& 0.71
& $1.76\times10^{-5}$ \\
\midrule
\multicolumn{7}{l}{\textit{Diffusion-enhanced neural operators}} \\
DM-FNO
& 0.34
& 0.64
& $8.79\times10^{-6}$
& 0.27
& 0.72
& $6.70\times10^{-6}$ \\
DM-UNO
& 0.35
& 0.64
& $9.22\times10^{-6}$
& 0.26
& 0.72
& $5.76\times10^{-6}$ \\
\midrule
\multicolumn{7}{l}{\textit{MENO models}} \\
MENO-FNO
& 0.21
& $\mathbf{0.76}$
& $\mathbf{8.62\times10^{-6}}$
& 0.09
& 0.92
& $6.25\times10^{-6}$ \\
MENO-UNO
& 0.21
& $\mathbf{0.76}$
& $9.06\times10^{-6}$
& $\mathbf{0.08}$
& $\mathbf{0.93}$
& $\mathbf{5.16\times10^{-6}}$ \\
\midrule
Relative improvement (\%)
& n/a
& 4.1
& 29.3
& 63.6
& 31.0
& 70.7 \\
\bottomrule
\end{tabular}
\end{table}

Table~\ref{tab:kf256_size_time} confirms that this improvement in turbulent
rollout quality is achieved without increasing the model size. The
diffusion-enhanced and MENO variants use the same neural operator and decoder
parameter budgets within each backbone: DM-FNO and MENO-FNO both use a
27.10M-parameter FNO with a 6.63M-parameter decoder, while DM-UNO and
MENO-UNO both use a 26.83M-parameter UNO with the same decoder size. This
matched-capacity comparison isolates the computational effect of replacing
multi-step diffusion sampling with the proposed single-step decoder. Under this
setting, MENO-FNO reduces the per-frame inference time from
$0.377\pm0.008$~s to $0.030\pm0.004$~s, and MENO-UNO reduces it from
$0.376\pm0.008$~s to $0.030\pm0.004$~s. Both correspond to approximately a
$12.5\times$ speedup. This efficiency gain is especially important for KF256
because turbulent rollouts require many autoregressive steps; reducing the
cost per frame directly lowers the cost of long-horizon prediction while
retaining the spectral and temporal advantages shown above.

\begin{table}[t]
\centering
\caption{KF256 model size and inference time for generative-enhanced neural
operators. Timings are measured per predicted frame for the full pipeline,
including the neural operator and the decoder.}
\label{tab:kf256_size_time}
\footnotesize
\setlength{\tabcolsep}{8pt}
\begin{tabular}{lcc}
\toprule
Model
& Parameters, NO + decoder
& Inference time per frame (s) \\
\midrule
\multicolumn{3}{l}{\textit{Diffusion-enhanced neural operators}} \\
DM-FNO
& 27.10M + 6.63M
& $0.377 \pm 0.008$ \\
DM-UNO
& 26.83M + 6.63M
& $0.376 \pm 0.008$ \\
\midrule
\multicolumn{3}{l}{\textit{MENO models}} \\
MENO-FNO
& 27.10M + 6.63M
& $\mathbf{0.030 \pm 0.004}$ \\
MENO-UNO
& 26.83M + 6.63M
& $\mathbf{0.030 \pm 0.004}$ \\
\bottomrule
\end{tabular}
\end{table}

\subsection{Active matter}
\label{Exp:am}

Finally, we consider an active matter system: a suspension of self-driven
rod-like particles with finite excluded volume immersed in a Stokes
fluid~\cite{maddu2024learning}. Unlike Cahn-Hilliard (dissipative,
gradient-flow) and Kolmogorov flow (forced, statistically stationary), active
matter is intrinsically non-equilibrium, with persistent injection of energy at
the particle scale producing heterogeneous, intermittent patterns. The
benchmark therefore tests whether a single-step generative decoder can recover
small-scale features that arise from long-range hydrodynamic coupling rather
than from a closed deterministic equation. We use the scalar concentration
field from the active-matter subset of \textit{The
Well}~\cite{ohana2024well,maddu2024learning} at $256\times256$ resolution,
following its standard data splits and evaluation protocol. The two resolution-transfer
settings are $32\rightarrow256$ and $64\rightarrow256$.

\begin{figure}[!h]
    \centering
    \includegraphics[width=0.95\linewidth]{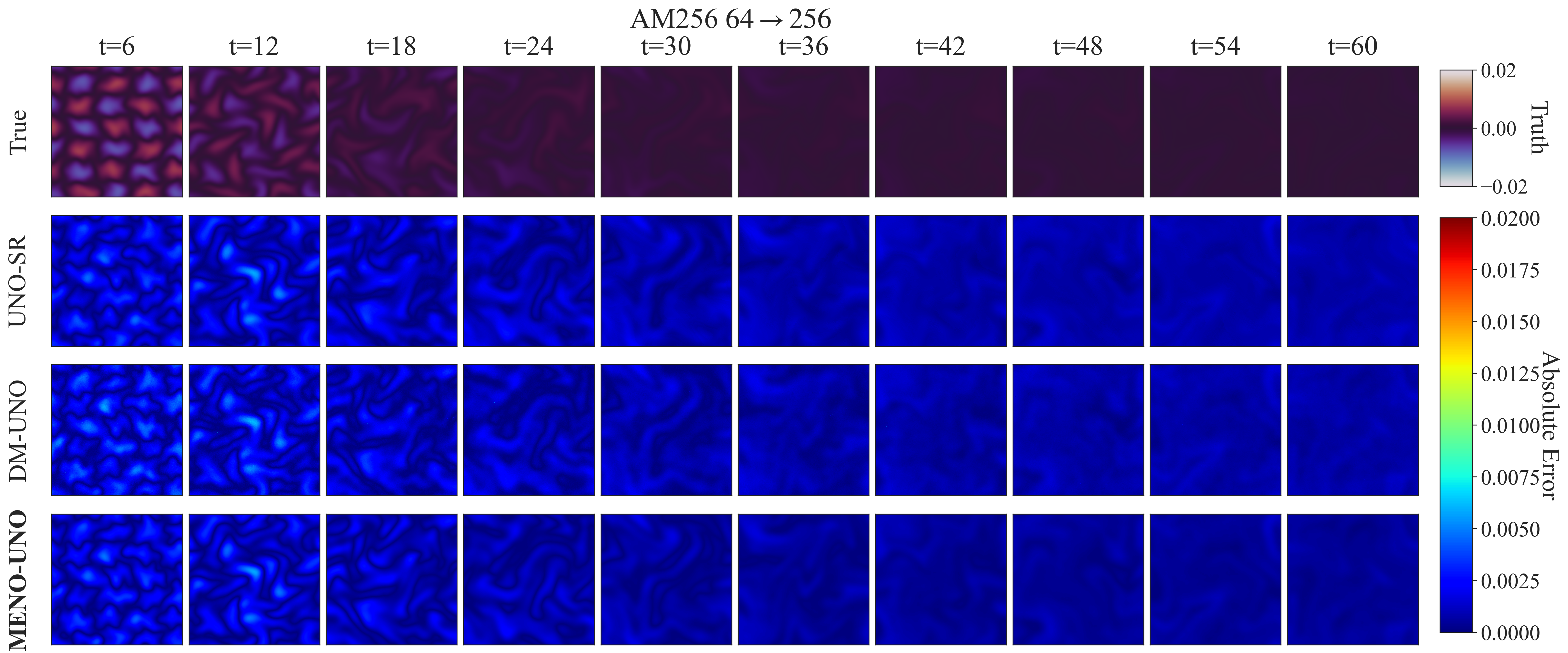}
    \caption{Snapshots along an uncurated AM256 trajectory for the
    $64\rightarrow256$ setting. The top row presents the high-resolution ground
    truth, and the next three rows visualize absolute errors for UNO-SR,
    MENO-UNO, and DM-UNO, respectively. Frames are normalized per frame to
    highlight error evolution since the mean concentration drifts over time.}
    \label{fig:am256_vis}
\end{figure}

\begin{figure}[!h]
    \centering
    \includegraphics[width=0.85\linewidth]{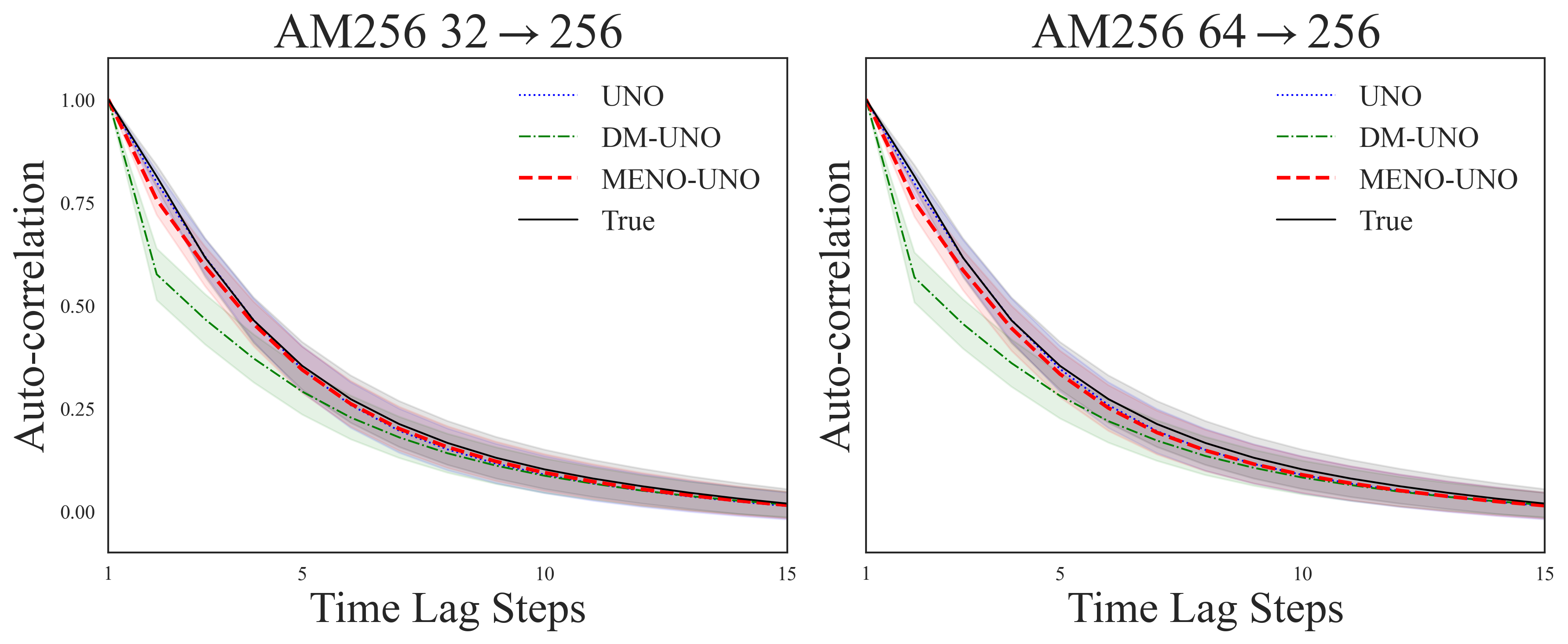}
    \caption{Temporal autocorrelation functions on AM256 across all resolution
    settings and model variants. Shaded bands denote the SEM over 25 test
    trajectories. The early-time decay is emphasized because it captures the
    dominant temporal coherence of the active concentration field.}
    \label{fig:am256_autocorr}
\end{figure}

Table~\ref{tab:am256_metrics} reports the AM256 results for UNO, DM-UNO, and
MENO-UNO. This dataset differs from the previous two because the concentration
field is generated by non-equilibrium active dynamics rather than by a closed
dissipative or forced Navier-Stokes equation. The relevant structures are
therefore heterogeneous and intermittent, and the surrogate must recover
small-scale concentration variations induced by hydrodynamic coupling and
particle-scale activity. In the $32\rightarrow256$ setting, MENO-UNO reduces
the relative $L_2$ error from $2.11\times10^{-3}$ to $1.02\times10^{-3}$, a
51.7\% improvement over the direct UNO baseline. It also improves SSIM from
0.72 to 0.74 and lowers PSDD from $1.09\times10^{-5}$ to
$1.02\times10^{-5}$. The larger gain in relative $L_2$ compared with SSIM
indicates that MENO substantially improves the amplitude and placement of
concentration features, while the overall visual structure is already partly
captured by the UNO backbone.

The $64\rightarrow256$ setting is more challenging to interpret because the
direct UNO baseline is already strong in relative $L_2$, with an error of
$2.05\times10^{-3}$. MENO still gives the best value on every metric, reducing
the error to $2.03\times10^{-3}$, increasing SSIM to 0.75, and reducing PSDD to
$1.08\times10^{-5}$. Although the relative $L_2$ gain is modest in this
setting, the consistent improvements in SSIM and PSDD show that MENO preserves
spatial organization and spectral content more reliably than direct
super-resolution. By contrast, DM-UNO degrades all three metrics in both
resolution settings, most notably reducing SSIM to 0.56. This suggests that
diffusion-based refinement is less effective for AM256: the stochastic
denoising process can introduce local texture that is not consistent with the
active concentration patterns.

Figure~\ref{fig:l2_and_spec}(c) shows the time-resolved behavior for the
$64\rightarrow256$ case. The relative $L_2$ error decreases over time for all
models, which reflects the progressive homogenization of the concentration
field rather than an improvement in predictive skill. Within this overall
trend, MENO remains below the diffusion-enhanced baseline and better matches
the high-frequency spectral tail of the ground truth. This is important for
active matter because small-scale concentration fluctuations are generated
continually by particle-scale activity; a model that over-smooths the field or
adds inconsistent high-frequency noise will miss the statistics of the active
state even if the averaged error remains small.

The qualitative examples in Figure~\ref{fig:am256_vis} are consistent with the
metric trends. Direct UNO-SR captures the broad concentration distribution but
leaves larger residual errors in localized regions where fine structures form
or dissipate. DM-UNO shows stronger discrepancies, consistent with the drop in
SSIM reported in Table~\ref{tab:am256_metrics}. MENO-UNO produces lower error
magnitudes across the rollout and better preserves the spatial arrangement of
the concentration field. Since AM256 frames are normalized per frame in the
visualization to account for drift in mean intensity, the comparison emphasizes
the evolving spatial error pattern rather than only the absolute field
magnitude.

\begin{table}[h]
\centering
\caption{AM256 model size and inference time for generative-enhanced neural
operators. Timings are measured per predicted frame for the full pipeline,
including the neural operator and the decoder.}
\label{tab:am256_size_time}
\footnotesize
\setlength{\tabcolsep}{8pt}
\begin{tabular}{lcc}
\toprule
Model
& Parameters, NO + decoder
& Inference time per frame (s) \\
\midrule
\multicolumn{3}{l}{\textit{Diffusion-enhanced neural operators}} \\
DM-UNO
& 26.83M + 6.63M
& $0.456 \pm 0.011$ \\
\midrule
\multicolumn{3}{l}{\textit{MENO models}} \\
MENO-UNO
& 26.83M + 6.63M
& $\mathbf{0.032 \pm 0.005}$ \\
\bottomrule
\end{tabular}
\end{table}

The efficiency comparison in Table~\ref{tab:am256_size_time} is particularly
important for AM256 because the active-matter rollout is evaluated over many
frames and the refinement step is applied at every predicted time. DM-UNO and
MENO-UNO have identical model sizes, using the same 26.83M-parameter UNO
backbone and the same 6.63M-parameter decoder. Hence, the runtime difference is
not due to parameter reduction, but to the fact that MENO replaces iterative
diffusion sampling with a single-step refinement. This reduces the per-frame
inference time from $0.456\pm0.011$~s for DM-UNO to $0.032\pm0.005$~s for
MENO-UNO, a $14.3\times$ speedup. Combined with the metric, spectral,
qualitative, and autocorrelation results, this shows that MENO is not only more
accurate than diffusion refinement on AM256 but also much more practical for
long non-equilibrium rollouts.

\begin{table}[!h]
\centering
\caption{AM256 prediction metrics for direct neural operator super-resolution,
diffusion-enhanced neural operators, and MENO. $\mathrm{R}L_2$ and SSIM are
computed over the first 10 frames, while PSDD is computed over the full
80-frame trajectories. The best value in each column is shown in bold.
Relative improvement is computed with respect to the corresponding direct
neural operator baseline.}
\label{tab:am256_metrics}
\footnotesize
\setlength{\tabcolsep}{4pt}
\begin{tabular}{lcccccc}
\toprule
& \multicolumn{3}{c}{$32\rightarrow256$}
& \multicolumn{3}{c}{$64\rightarrow256$} \\
\cmidrule(lr){2-4} \cmidrule(lr){5-7}
Model
& $\mathrm{R}L_2 \downarrow$
& SSIM $\uparrow$
& PSDD $\downarrow$
& $\mathrm{R}L_2 \downarrow$
& SSIM $\uparrow$
& PSDD $\downarrow$ \\
\midrule
\multicolumn{7}{l}{\textit{Neural operators without enhancement}} \\
UNO
& $2.11\times10^{-3}$
& 0.72
& $1.09\times10^{-5}$
& $2.05\times10^{-3}$
& 0.73
& $1.12\times10^{-5}$ \\
\midrule
\multicolumn{7}{l}{\textit{Diffusion-enhanced neural operators}} \\
DM-UNO
& $2.17\times10^{-3}$
& 0.56
& $1.24\times10^{-5}$
& $2.14\times10^{-3}$
& 0.56
& $1.30\times10^{-5}$ \\
\midrule
\multicolumn{7}{l}{\textit{MENO models}} \\
MENO-UNO
& $\mathbf{1.02\times10^{-3}}$
& $\mathbf{0.74}$
& $\mathbf{1.02\times10^{-5}}$
& $\mathbf{2.03\times10^{-3}}$
& $\mathbf{0.75}$
& $\mathbf{1.08\times10^{-5}}$ \\
\midrule
Relative improvement (\%)
& 51.7
& 2.8
& 6.4
& 1.0
& 2.7
& 3.6 \\
\bottomrule
\end{tabular}
\end{table}

For the active matter system, temporal autocorrelation provides a complementary
test of whether the surrogate captures persistent non-equilibrium dynamics
rather than only instantaneous spatial structure. We compute the
autocorrelation of the concentration fields along the time axis using the same
Fourier-transform estimator described in Appendix~\ref{appendix:metrics}.
Figure~\ref{fig:am256_autocorr} shows that direct UNO and DM-UNO decorrelate at
rates that deviate from the ground truth, showing that instantaneous field
reconstruction does not guarantee correct temporal statistics. MENO more
closely follows the ground-truth decay in both resolution settings, indicating
that the refined concentration fields retain the memory of the underlying
active dynamics. The AM256 results therefore demonstrate that MENO is not
limited to PDE systems with explicit dissipative or turbulent structure: it
also improves non-equilibrium active matter rollouts, where the main challenge
is to recover intermittent small-scale features while preserving temporal
coherence.

The three benchmarks probe complementary regimes: dissipative
gradient flow with sharp diffuse interfaces (Cahn-Hilliard), forced
statistically stationary turbulence with a broadband spectrum (Kolmogorov
flow), and non-equilibrium active matter driven by particle-scale activity
(AM256). On all three, MENO is the only method that simultaneously (i) lowers relative $L_2$ in well-resolved settings, (ii) reduces PSDD relative to both direct neural operator super-resolution and diffusion-enhanced refinement, and (iii) tracks the system-specific physical diagnostic, namely the Ginzburg-Landau free-energy decay for Cahn-Hilliard, the high-wavenumber spectral tail for Kolmogorov flow, and the temporal autocorrelation in all cases. These improvements are obtained at matched neural operator and decoder parameter budgets, with per-frame inference cost reduced from $0.18$-$0.46$~s for diffusion-enhanced refinement to $0.03$~s for MENO. The
consistency of these gains across three physically distinct systems indicates that the improvements are intrinsic to the single-step i-MF decoder rather than tied to a particular dataset or physical regime.

\section{Conclusion}
\label{sec:conclusion}

We introduced MeanFlow-Enhanced Neural Operators, a two-stage framework that overcomes a known failure mode of neural operators: their inability to recover fine-scale structure when evaluated above the training resolution. A neural operator first learns the coarse-grained temporal dynamics on a low-resolution grid, and a single-step generative decoder then restores high-resolution structure from the upsampled rollout in one network evaluation. We build the decoder on the improved MeanFlow objective rather than the original MeanFlow: the improved formulation rewrites the self-consistency identity as a standard regression on the instantaneous velocity, eliminating the diverging training loss of the original formulation and yielding lower relative $L_2$ error and power-spectrum density discrepancy at matched compute. On Cahn-Hilliard phase-field dynamics, two-dimensional Kolmogorov flow, and active matter, the framework outperforms both direct neural operator super-resolution and diffusion-enhanced refinement, reducing the power-spectrum density discrepancy by up to 70\%, reproducing the Cahn-Hilliard free-energy decay and the long-time autocorrelation curves of all three systems, and running $5.7\times$ (Cahn-Hilliard phase field), $12.5\times$ (Kolmogorov flow), and $14.3\times$ (active matter) faster than denoising-diffusion implicit refinement at matched parameter budgets.

Three limitations remain. The optimal noise level $\tau$ in the improved MeanFlow decoder depends on the resolution gap, so accuracy degrades when the autoregressive operator drifts far from its training distribution. Our experiments also fix the governing-equation parameters, leaving generalisation across Reynolds numbers, material parameters, and boundary conditions untested. Finally, the proposed framework enforces no hard physical constraints; it learns mass conservation and incompressibility only statistically, from data.

A natural next step is to scale the framework to fully three-dimensional, multi-physics systems, where iterative diffusion refinement becomes prohibitive and the speedup from a single-step decoder is most valuable. More broadly, this work paves the way for combining generative modelling with neural operators as a general route to fast, physically consistent surrogates for high-resolution dynamical systems.

\section*{Acknowledgments}
The authors acknowledge the use of compute resources provided by the
Isambard-AI National AI Research Resource (AIRR). Isambard-AI is operated by
the University of Bristol and funded by the UK Government's Department for
Science, Innovation and Technology (DSIT) through UK Research and Innovation
(UKRI).

\bibliographystyle{elsarticle-num}
\bibliography{ref}

\clearpage
\appendix
\section{DDIM-Accelerated Sampling for the Diffusion Decoder}\label{appendix:dmno}
The diffusion-decoder baseline is trained with the standard DDPM objective; the forward process and the corresponding noise-prediction loss are given in the main text (Eq.~\ref{eq:ddpm_forward}) and follow~\cite{ho2020denoising}. At inference time, the reverse Markov chain
\begin{equation}
\mathbf{x}_{t-1} = \frac{1}{\sqrt{\alpha_t}}\!\left(\mathbf{x}_t - \frac{\beta_t}{\sqrt{1 - \bar{\alpha}_t}}\,\boldsymbol{\epsilon}_\theta(\mathbf{x}_t, t)\right) + \boldsymbol{\epsilon} \sqrt{\tfrac{1 - \bar{\alpha}_{t-1}}{1 - \bar{\alpha}_t}\, \beta_t},\quad \boldsymbol{\epsilon}\sim\mathcal{N}(0,I),
\end{equation}
requires $T$ network evaluations (typically $T=1000$). To make the diffusion baseline competitive in inference cost, we adopt DDIM~\cite{song2020denoising}, which uses a non-Markovian parametrization to skip intermediate steps along a deterministic trajectory. The procedure used in our experiments is summarized in Algorithm~\ref{alg:ddim_sampling}.
\begin{algorithm}[H]
\caption{DDIM-Enhanced Neural Operator (inference)}
\label{alg:ddim_sampling}
\begin{algorithmic}[1]
\REQUIRE DDPM noise predictor $\boldsymbol{\epsilon}_\theta$, schedule $\{\bar\alpha_t\}_{t=0}^{T}$, sub-steps $0=\tau_0<\cdots<\tau_K$, low-resolution initial state $a_0^{\text{LR}}$, trained neural operator $\mathcal{G}_{\phi^*}$, horizon $T$, uniform upsampler $U(\cdot)$
\FOR{$t=1,\dots,T$}
  \STATE $\tilde{\mathbf{a}}^\text{LR}_t \leftarrow \mathcal{G}_{\phi^*}(\tilde{\mathbf{a}}^\text{LR}_{t-1})$
  \STATE $s \gets \tau_{k}$,\; $u \gets \tau_{k-1}$
  \STATE $\boldsymbol{\epsilon}_t \gets \boldsymbol{\epsilon}_\theta(U(\tilde{\mathbf{a}}^\text{LR}_t),s)$
  \STATE $\tilde{\mathbf{x}}_{0,t} \gets \dfrac{U(\tilde{\mathbf{a}}^\text{LR}_t) - \sqrt{1-\bar\alpha_s}\,\boldsymbol{\epsilon}_t}{\sqrt{\bar\alpha_s}}$
  \STATE $\hat{\mathbf{x}}_{0,t} \gets \sqrt{\bar\alpha_u}\,\tilde{\mathbf{x}}_{0,t} + \sqrt{1-\bar\alpha_u}\,\boldsymbol{\epsilon}$
\ENDFOR
\end{algorithmic}
\textbf{Return:} High-resolution forecast $\{\hat{\mathbf{x}}_t\}_{t=1}^T$.
\end{algorithm}

\section{Impact of Denoising Settings for MENOs and Diffusion Models}\label{appendix:denoise}
\subsection{Diffusion Models}
\begin{figure*}[h]
    \centering    \includegraphics[width=0.5\linewidth]{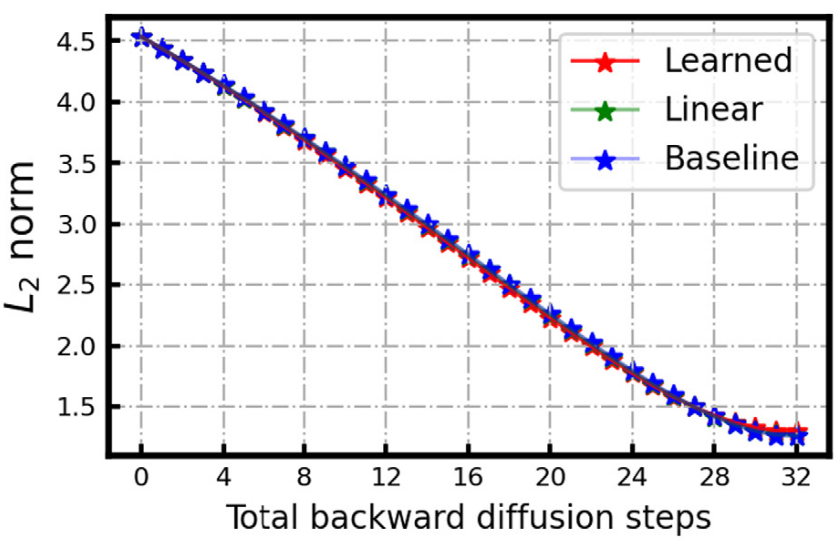}
    \caption{Image adapted from~\cite{shu2023physics}, showing the reconstruction error trends for the $32 \rightarrow 256$ generative refinement task. The legend compares three reconstruction strategies; in this work, we focus on the \textit{Baseline} method.}
    \label{fig:kf_l2}
\end{figure*}

Shu et al.~\cite{shu2023physics} extensively study diffusion-based generative decoders for KF256, using the procedure summarized in Appendix~\ref{appendix:dmno}. The key observation we use is shown in Figure~\ref{fig:kf_l2}: when the low-resolution input is ground truth (unlike our setting where low-resolution states may be predicted by neural operators), the reconstruction error decreases monotonically as the number of DDIM steps increases. Guided by this efficiency-accuracy trade-off, we use 20 DDIM steps with a noise level of 400, which provides a strong diffusion baseline and avoids overstating the gains of MENO.

\subsection{MENO Noise Strength}
\begin{figure*}[h]
    \centering    \includegraphics[width=0.7\linewidth]{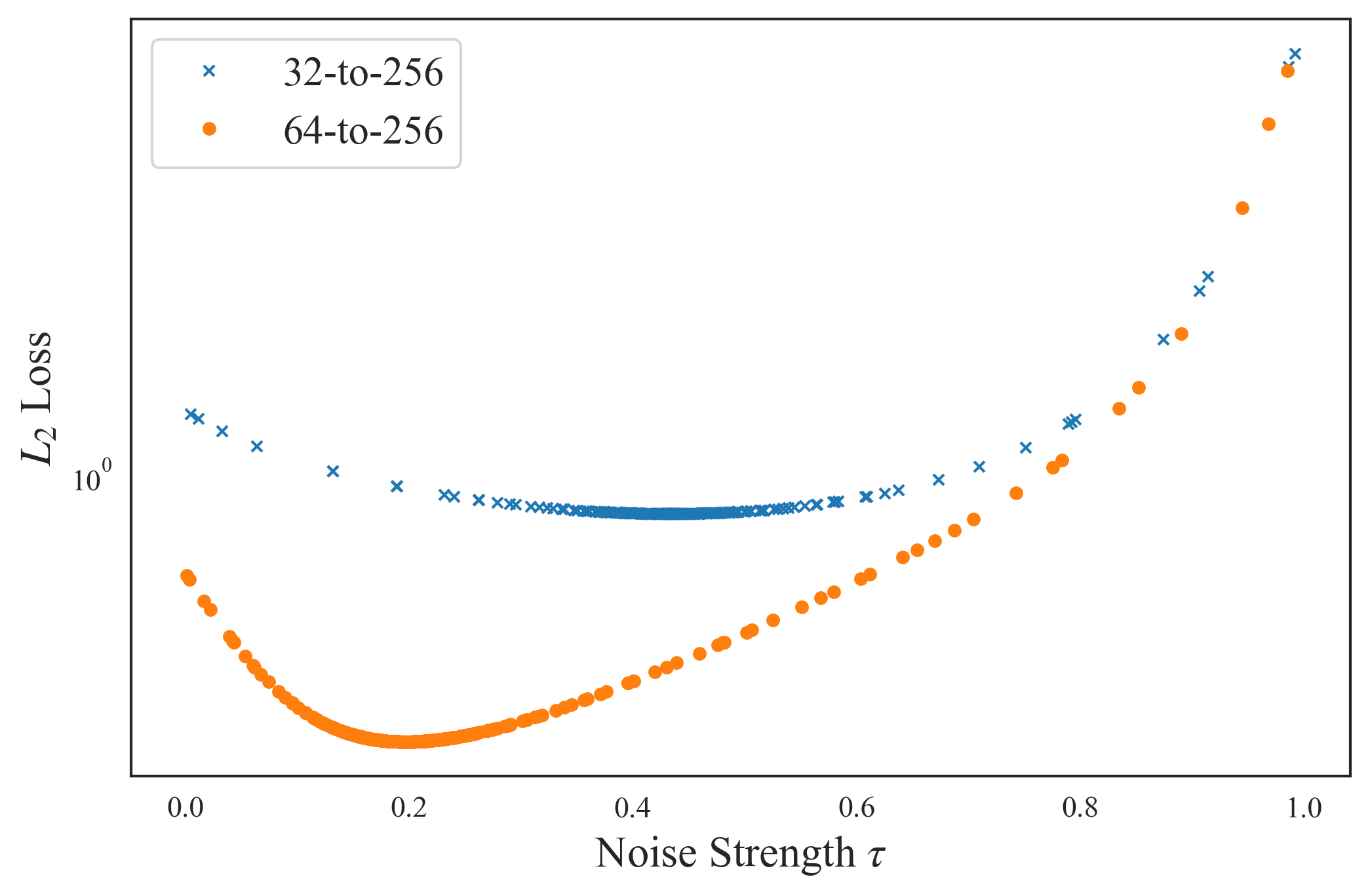}
    \caption{Absolute $L_2$ loss of MENO decoder models on the KF256 dataset for the $32 \rightarrow 256$ and $64 \rightarrow 256$ generative refinement tasks.}
    \label{fig:kf_mf_l2}
\end{figure*}

Since i-MF is an intrinsic one-step method, the only tuning parameter is the noise level. We perform a weighted stochastic search over $\tau \in (0,1]$ and record the reconstruction $L_2$ loss using \emph{Optuna}, with results shown in Figure~\ref{fig:kf_mf_l2}. To avoid data leakage, this study uses ground-truth low-resolution inputs rather than neural operator predictions. We evaluate both $32 \rightarrow 256$ and $64 \rightarrow 256$ generative refinement tasks. 
While the overall trend is similar, the optimal noise levels differ, with $\tau \approx 0.432$ for $32 \rightarrow 256$ and $\tau \approx 0.197$ for $64 \rightarrow 256$. Intuitively, lower-resolution inputs require stronger noise to better match the distribution of intermediate states, whereas higher-resolution inputs benefit from weaker noise. This provides a qualitative explanation of i-MF decoder: it implicitly relies on distributional similarity between corrupted low-resolution states and true intermediate states. Consequently, performance is expected to degrade when the low-resolution input deviates substantially from the ground truth.

\subsection{Distributional Drift}
\begin{figure*}[h]
    \centering    \includegraphics[width=0.7\linewidth]{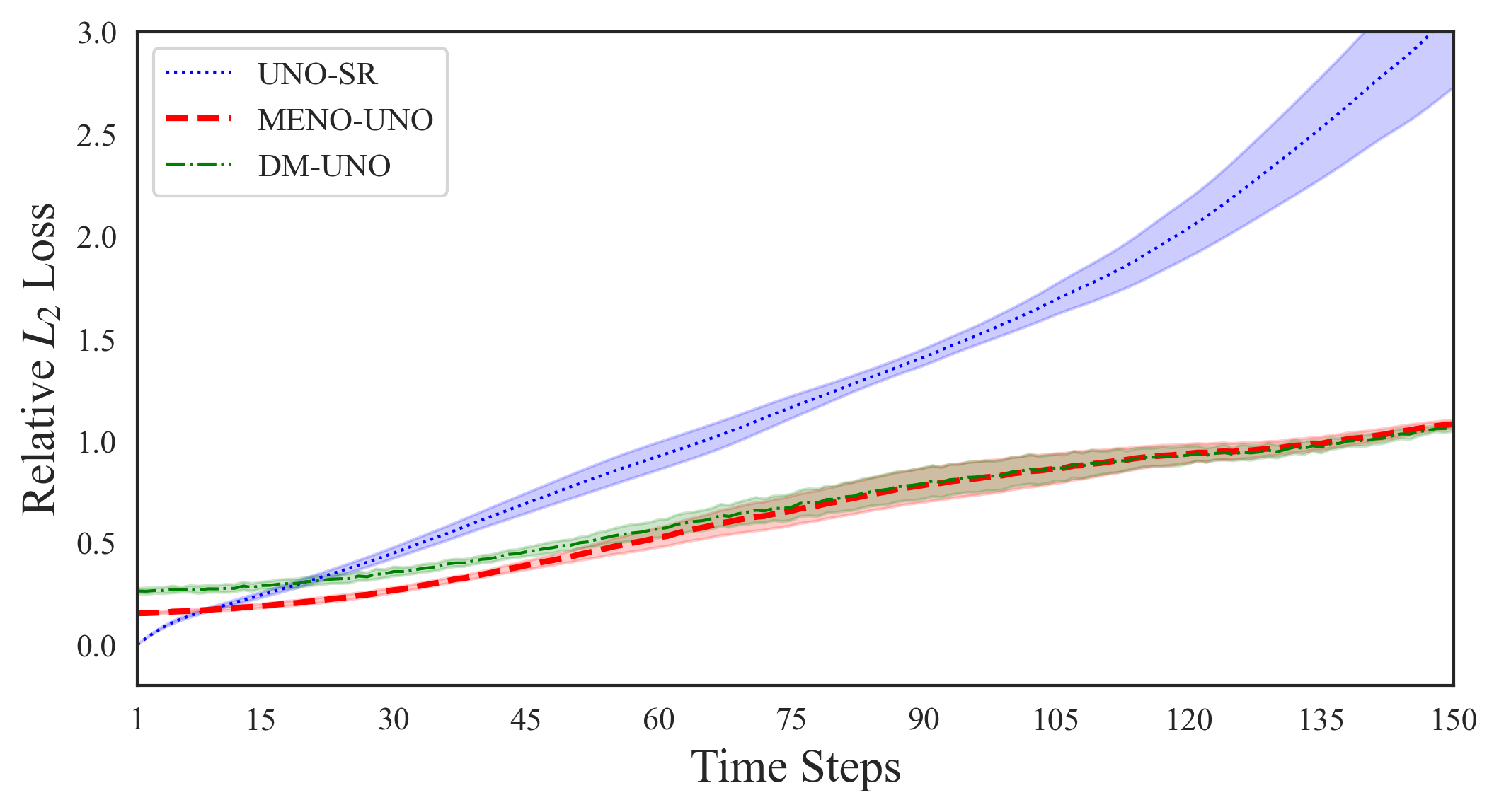}
    \caption{Long range relative $L_2$ loss of all UNO-based models on the KF256 dataset for the $32 \rightarrow 256$ generative refinement tasks. Shaded regions denote the SEM computed over 8 test trajectories.}
    \label{fig:kf_drift}
\end{figure*}

Although a fully quantitative analysis is beyond the scope of this study, we can qualitatively illustrate the effect of distributional drift in the low-resolution latent. During autoregressive rollouts, neural operators inevitably accumulate errors, causing later frames to deviate increasingly from the ground truth. This behavior is reflected by the long-range growth of relative $L_2$ error in Figure~\ref{fig:kf_drift}.

 While these later frames lie outside the prediction range, generative refinement can still substantially reduce error with respect to the ground truth, indicating that the decoder remains effective under moderate, non-severe drift.

\subsection{Comparison between Improved MeanFlow and the Original MeanFlow}

\begin{figure*}[h]
    \centering
    \includegraphics[width=1.\linewidth]{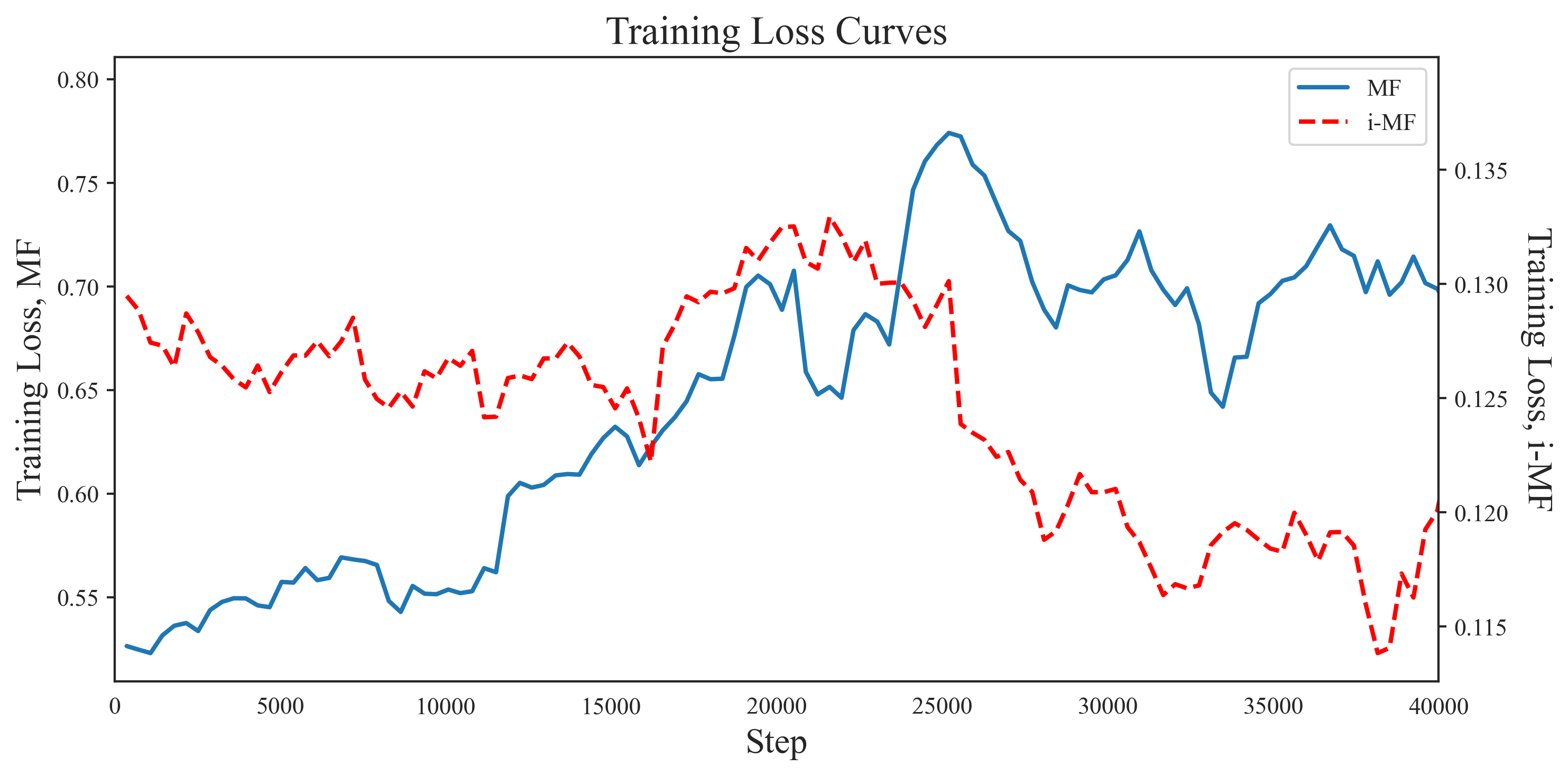}
    \caption{Training loss curves of i-MF and the original MF on the KF256 dataset, shown with moving-average smoothing. The loss of the original MeanFlow increases as training proceeds, consistent with the behavior reported in Figure~3 of~\cite{geng2025improved}.}
    \label{fig:imf_vs_mf}
\end{figure*}

Geng et al.~\citep{geng2025improved} reported that the training loss of the original MeanFlow can increase even as generation quality improves. We observe the same phenomenon in our experiments, as shown in Figure~\ref{fig:imf_vs_mf}. This behavior provides a direct indication of the training instability associated with the original MeanFlow formulation. To further quantify the effect of this instability, we train both variants for the same number of epochs on the KF256 dataset and report the results in Table~\ref{tab:kf256_imf_mf_ablation}.

\begin{table}[t]
\centering
\caption{Ablation study on the KF256 dataset comparing i-MF- and MF-based variants of MENO.
All results are reported for the $32\rightarrow256$ setting. }
\label{tab:kf256_imf_mf_ablation}
\footnotesize
\setlength{\tabcolsep}{8pt}
\begin{tabular}{lccc}
\toprule
Model
& $\mathrm{R}L_2 \downarrow$
& SSIM $\uparrow$
& PSDD $\downarrow$ \\
\midrule
\multicolumn{4}{l}{\textit{i-MF variants}} \\
i-MF:MENO-FNO
& $0.21$
& $0.76$
& $8.62\times10^{-6}$ \\
i-MF:MENO-UNO
& $0.21$
& $0.76$
& $9.06\times10^{-6}$ \\
\midrule
\multicolumn{4}{l}{\textit{MF variants}} \\
MF:MENO-FNO
& 0.23
& 0.75
& $8.96\times10^{-6}$ \\
MF:MENO-UNO
& 0.25
& 0.74
& $9.21\times10^{-6}$ \\
\bottomrule
\end{tabular}
\end{table}

\subsection{Impact of Generation Randomness}\label{appendix:randomness}
In this section, we shall illustrate how the stochastic nature of generative models affect the metric values. To investigate the impact of random seed, we use 100 randomly generated integers for the generative decoder and see how uncertainty arises across different runs on the KF256 tasks. The result is summarised in Table \ref{tab:kf256_metrics_rand}. This small errors confirms that the metric values, up to the digits we report in the main body, is not affected by the randomness of generative models.

\begin{table}[!h]
\centering
\caption{KF256 $32\rightarrow256$ metrics for DM-enhanced NOs, and MENOs. $\mathrm{R}L_2$ and SSIM are computed over the first 20 frames, while PSDD is computed over the full trajectories (180 frames). Uncertainties are computed over 100 runs with different random seeds. We report values to four significant figures; uncertainties smaller than this precision are rounded and reported as $\pm 1$ in the last digit.
}
\label{tab:kf256_metrics_rand}
\setlength{\tabcolsep}{0.3em}
\begin{tabular}{lccc ccc}
\toprule
& \multicolumn{3}{c}{\textbf{32 $\rightarrow$ 256}} & \multicolumn{3}{c}{\textbf{64 $\rightarrow$ 256}} \\
\cmidrule(lr){2-4} \cmidrule(lr){5-7}
\textbf{Model} & $\mathrm{R}L_2 \downarrow$ & SSIM $\uparrow$ & PSDD $\downarrow$
              & $\mathrm{R}L_2 \downarrow$ & SSIM $\uparrow$ & PSDD $\downarrow$ \\
\midrule
DM-UNO   & 0.3541(4) & 0.6383(2)  & $(9.221\pm0.001)\times 10^{-6}$ & 0.2581(1)  & 0.7178(3) & $(5.764\pm0.001)\times 10^{-6}$ 
\\
MENO-UNO & 0.2131(1) & 0.7612(1)  & $(9.058\pm0.001)\times 10^{-6}$ & 0.0813(1)  & 0.9321(2) & $(5.161\pm0.001)\times 10^{-6}$ 
\\
\bottomrule
\end{tabular}
\end{table}

\clearpage

\section{Dataset preparation details}\label{appendix: datasets and metrics}
The governing equations and high-level dataset descriptions are given in Section~\ref{section:experiments}. Here we summarize additional implementation details used to generate the supervised samples.

\paragraph{PF100 (Cahn-Hilliard)} Each of the 300 independent runs is performed in COMSOL Multiphysics on a uniform $100\times 100$ Cartesian grid over $\Omega=[0,1]^2$, starting from the homogeneous initial condition $\phi(\mathbf{x},0)=0$ with run-specific random seeds controlling the stochastic perturbations. Each trajectory contains 25 consecutive frames at a fixed sampling interval $\Delta t$~\cite{xue2025equivariant}. From each trajectory we extract overlapping $(n_{\text{in}},n_{\text{out}})$ input-output windows by sliding along the time axis. The dataset is split 80\%/10\%/10\% into train/validation/test.

\paragraph{KF256 (Kolmogorov flow)} The 70 trajectories at $\mathrm{Re}=1000$ are generated with the JAX-CFD community code~\cite{Kochkov2021-ML-CFD} on a doubly periodic $256\times 256$ grid. Each trajectory contains 180 vorticity-field frames, from which overlapping windows are extracted as for PF100. Dataset URL: \url{https://figshare.com/ndownloader/files/39181919}.

\paragraph{AM256 (Active matter)} We use the scalar concentration channel of the active-matter subset of \textit{The Well}~\cite{ohana2024well, maddu2024learning} at $256\times 256$, following the dataset's standard splits and evaluation protocol. Dataset URL: \url{https://polymathic-ai.org/the_well/datasets/active_matter/}.
\clearpage

\section{Implementation details of the evaluation metrics}\label{appendix:metrics}
The $\mathrm{R}L_2$, SSIM, and PSDD definitions are stated in Section~\ref{subsec:metrics} (Eqs.~\ref{eq:rl2}-\ref{eq:psdd}). Below we record the implementation choices used to make these metrics numerically robust on unnormalized scientific fields.

\paragraph{SSIM: Gaussian-window statistics} Local means $\mu_x,\mu_y$ and (biased) local variances/covariance $\sigma_x^2,\sigma_y^2,\sigma_{xy}$ are estimated through a Gaussian smoothing operator $\mathcal{H}$ implemented as a separable 2D convolution with a normalized 1D Gaussian kernel of size $w\times w$ and same-padding, so spatial resolution is preserved. The stability constants $C_1=(K_1 L)^2$ and $C_2=(K_2 L)^2$ use the standard choices $K_1=0.01$, $K_2=0.03$, and the data range $L = \max(x)-\min(x)$ is computed \emph{per sample} from the reference field. When $L=0$ (constant reference), we return $\overline{\mathrm{SSIM}}=1$ if $x$ and $y$ agree within numerical tolerance and $0$ otherwise. For batched inputs of shape $(B,H,W,1)$ we compute SSIM independently per sample and report the batch average.

\paragraph{PSDD: overflow-safe normalized PSD} Before the 2D DFT we subtract the per-sample mean of each field so that the DC component does not dominate the spectrum. To avoid overflow when squaring high-dynamic-range scientific fields, we rescale the complex FFT output by the per-sample maximum of $\{|\Re\widetilde F|,|\Im\widetilde F|\}$ before computing the power $P=\Re(\widetilde F)^2+\Im(\widetilde F)^2$. Each spectrum is then converted into a discrete probability distribution $\widehat P = P/\sum P$, removing sensitivity to amplitude and isolating the relative allocation of spectral energy used by Eq.~\ref{eq:psdd}.

\paragraph{autocorrelation function} For each batch element $b$ and spatial location $(h,w)$ we form the time series $x_{b,t}(h,w)$ for $t=0,\dots,T-1$, de-mean along $t$, and zero-pad to length $n_{\mathrm{fft}}\ge 2T-1$ to obtain the linear (non-circular) autocorrelation via the Wiener-Khinchin identity, $r_b(\ell;h,w)=\mathcal{F}^{-1}(|X_b|^2)[\ell]$. The auto-covariance estimate uses the unbiased normalization $\hat\gamma_b(\ell)=r_b(\ell)/(T-\ell)$, and the autocorrelation is reported as $\hat\rho_b(\ell)=\hat\gamma_b(\ell)/(\hat\gamma_b(0)+\varepsilon)$, averaged across the batch with the SEM as uncertainty band.
\clearpage

\section{Hyperparameters}\label{appendix:hyperparameters}

In this section, we give all the relevant experiment settings for reproducibility. Inference hyperparameters are discussed in Appendix \ref{appendix:denoise}. 

\subsection{PF100 Experiment Setting}
The neural operator architectures are summarized in Table~\ref{tab:pf100_nos} for the $20 \rightarrow 100$ and $50 \rightarrow 100$ tasks. Hyperparameters are chosen so that the two configurations have matched parameter counts, with comparable depth and a similar number of Fourier modes. All neural operators are trained by Adam optimizers with learning rate $0.0001$ for 1500 epochs. 

\begin{table}[h]
\centering
\small
\resizebox{0.85\textwidth}{!}{
\begin{tabular}{l|c|c}
\hline
\textbf{Parameter Name} & \textbf{FNO Configuration} & \textbf{UNO Configuration} \\
\hline
channel\_mlp\_dropout & 0 & 0 \\
channel\_mlp\_expansion & 0.5 & 0.5 \\
channel\_mlp\_skip & soft-gating & linear \\
factorization & tucker & tucker \\
fno\_skip & linear & linear \\
hidden\_channels & 78 & 128 \\
horizontal\_skip &  & linear \\
implementation & factorized & factorized \\
in\_channels & 1 & 1 \\
lifting\_channel\_ratio & 2 &  \\
lifting\_channels &  & 256 \\
n\_layers & 7 & 7 \\
n\_modes & (16,16) &  \\
out\_channels & 1 & 1 \\
projection\_channel\_ratio & 2 &  \\
projection\_channels &  & 64 \\
rank & 1.0 & 1.0 \\
uno\_n\_modes &  &
(32,32),(16,16),(8,8),(4,4),(8,8),(16,16),(32,32) \\
uno\_out\_channels &  & 32,64,64,128,64,64,32 \\
uno\_scalings &  &
(1,1),(0.5,0.5),(0.5,0.5),(1,1),(2,2),(2,2),(1,1) \\
\hline
\end{tabular}}
\caption{FNO and UNO architectures used in all tasks for PF100 dataset.}
\label{tab:pf100_nos}
\end{table}

The diffusion model setting and the neural network backbone architecture are shown in Table \ref{tab:pf100_backbone}. MF models share the same backbone model. All decoder modules are trained with Adam optimizers, learning rate $0.0001$ and weight decay $0.0001$, for 150 epochs. 

\begin{table}[t]
\centering
\begin{tabular}{p{0.3\linewidth} p{0.3\linewidth}}
\hline
\textbf{UNet hyperparameters} & \textbf{Model setting hyperparameters} \\
\hline
\begin{tabular}{@{}ll@{}}
attention\_resolutions: & (16) \\
channel\_multipliers: & (1, 1, 2) \\
latent\_dims: & 32 \\
num\_res\_blocks: & 3 \\
type: & UNet \\
\end{tabular}
&
\begin{tabular}{@{}ll@{}}
beta\_end: & 0.02 \\
beta\_start: & 0.0001 \\
diffusion\_steps: & 1000 \\
\end{tabular}
\\
\hline
\end{tabular}
\caption{Diffusion model setting for the PF100 experiment.}
\label{tab:pf100_backbone}
\end{table}

\subsection{KF256 Experiment Setting}
The neural operator architectures are summarized in Table~\ref{tab:kf256_nos} for the $64 \rightarrow 256$ and $32 \rightarrow 256$ tasks. Hyperparameters are chosen so that the two configurations have matched parameter counts, with comparable depth and a similar number of Fourier modes.

\begin{table}[h]
\centering
\small
\resizebox{0.85\textwidth}{!}{
\begin{tabular}{l|c|c}
\hline
\textbf{Parameter Name} & \textbf{FNO Configuration} & \textbf{UNO Configuration} \\
\hline
channel\_mlp\_dropout & 0 & 0 \\
channel\_mlp\_expansion & 2 & 2 \\
channel\_mlp\_skip & soft-gating & linear \\
factorization & tucker & tucker \\
fno\_skip & linear & linear \\
hidden\_channels & 84 & 64 \\
horizontal\_skip &  & linear \\
implementation & factorized & factorized \\
in\_channels & 1 & 1 \\
lifting\_channel\_ratio & 2 &  \\
lifting\_channels &  & 64 \\
n\_layers & 7 & 7 \\
n\_modes & (32,32) &  \\
out\_channels & 1 & 1 \\
projection\_channel\_ratio & 1 &  \\
projection\_channels &  & 64 \\
rank & 1.0 & 1.0 \\
uno\_n\_modes &  &
(64,64),(32,32),(32,32),(8,8),(4,4),(8,8),(32,32),(32,32),(64,64) \\
uno\_out\_channels &  & 32, 64, 64, 128, 64, 64, 32 \\
uno\_scalings &  &
(1,1),(1,1),(0.5,0.5),(1,1),(2,2),(1,1),(1,1) \\
\hline
\end{tabular}}
\caption{FNO and UNO architectures used in all tasks for KF256 dataset.}
\label{tab:kf256_nos}
\end{table}

The diffusion model setting and the neural network backbone architecture are shown in Table \ref{tab:kf256_backbone}. MF models share the same backbone model. 

\begin{table}[t]
\centering
\begin{tabular}{p{0.3\linewidth} p{0.3\linewidth}}
\hline
\textbf{UNet hyperparameters} & \textbf{Model setting hyperparameters} \\
\hline
\begin{tabular}{@{}ll@{}}
attention\_resolutions: & (16) \\
channel\_multipliers: & (1, 1, 1, 2) \\
latent\_dims: & 64 \\
num\_res\_blocks: & 3 \\
type: & UNet \\
\end{tabular}
&
\begin{tabular}{@{}ll@{}}
beta\_end: & 0.02 \\
beta\_start: & 0.0001 \\
diffusion\_steps: & 1000 \\
\end{tabular}
\\
\hline
\end{tabular}
\caption{Diffusion model setting for the PF100 experiment.}
\label{tab:kf256_backbone}
\end{table}

\subsection{AM256 Experiment Setting}
All neural operator models on AM256 are trained autoregressively using a five-step input window to predict the next frame. All other experimental settings for AM256 follow those used for KF256.
\clearpage

\end{document}